\documentclass[10pt,journal]{IEEEtran}

\usepackage{graphicx}
\usepackage{epsfig}
\usepackage{amsmath}
\usepackage{amssymb}
\usepackage{multirow}
\usepackage{longtable}
\usepackage{floatrow}
\usepackage{subfig}
\usepackage{threeparttable}
\usepackage{hyperref}
\usepackage{sidecap}

\newcommand{\x}{\mathbf{x}} 
\newcommand{\y}{\mathbf{y}} 
\newcommand{\s}{\mathbf{s}} 
\newcommand{\e}{\mathbf{e}} 
\newcommand{\se}{\,\mathcal{S}}
\newcommand{\un}{\,\mathcal{U}}
\newcommand{\rn}{\mathrm{N}}
\newcommand{\rs}{\mathrm{S}}
\newcommand{\ru}{\mathrm{U}}
 
\newcommand{\X}{\mathbf{X}} 
\newcommand{\W}{\mathbf{W}} 
\newcommand{\p}{\mathbf{p}}

\def\SR#1{{\color{blue} {{#1}}}}

\begin{document}

\title{A Unified approach for Conventional Zero-shot, Generalized Zero-shot and Few-shot Learning}


\author{Shafin Rahman, 
        Salman H. Khan 
        and Fatih Porikli 
}

\markboth{Journal of \LaTeX\ Class Files,~Vol.~14, No.~8, August~2017}%
{Shell \MakeLowercase{\textit{et al.}}: Bare Demo of IEEEtran.cls for IEEE Journals}

\maketitle

\begin{abstract}
Prevalent techniques in zero-shot learning do not generalize well to other related problem scenarios. Here, we present a unified approach for conventional zero-shot, generalized zero-shot and few-shot learning problems. Our approach is based on a novel Class Adapting Principal Directions (CAPD) concept that allows multiple embeddings of image features into a semantic space. Given an image, our method produces one principal direction for each seen class. Then, it learns how to combine these directions to obtain the principal direction for each unseen class such that the CAPD of the test image is aligned with the semantic embedding of the true class, and opposite to the other classes. This allows efficient and class-adaptive information transfer from seen to unseen classes. In addition, we propose an automatic process for selection of the most useful seen classes for each unseen class to achieve robustness in zero-shot learning. Our method can update the unseen CAPD taking the advantages of few unseen images to work in a few-shot learning scenario. Furthermore, our method can generalize the seen CAPDs by estimating seen-unseen diversity that significantly improves the performance of generalized zero-shot learning. Our extensive evaluations demonstrate that the proposed approach consistently achieves superior performance in zero-shot, generalized zero-shot and few/one-shot learning problems.
\end{abstract}


\IEEEpeerreviewmaketitle


\section{Introduction}

Being one of the most fundamental tasks in visual understanding, object classification has long been the focus of attention in computer vision. Recently, significant advances have been reported, in particular for supervised learning using deep learning based techniques that are driven by the emergence of large-scale annotated datasets, fast computational platforms, and efficient optimization methods \cite{Vgg_arXiv_2014,GNet_CVPR_2015}. 

\begin{figure*}[th]
  \centering
  \subfloat[Input image from unseen class Leopard]{\includegraphics[width=0.4\textwidth,trim={3.4cm 2cm 1.8cm 2.2cm},clip]{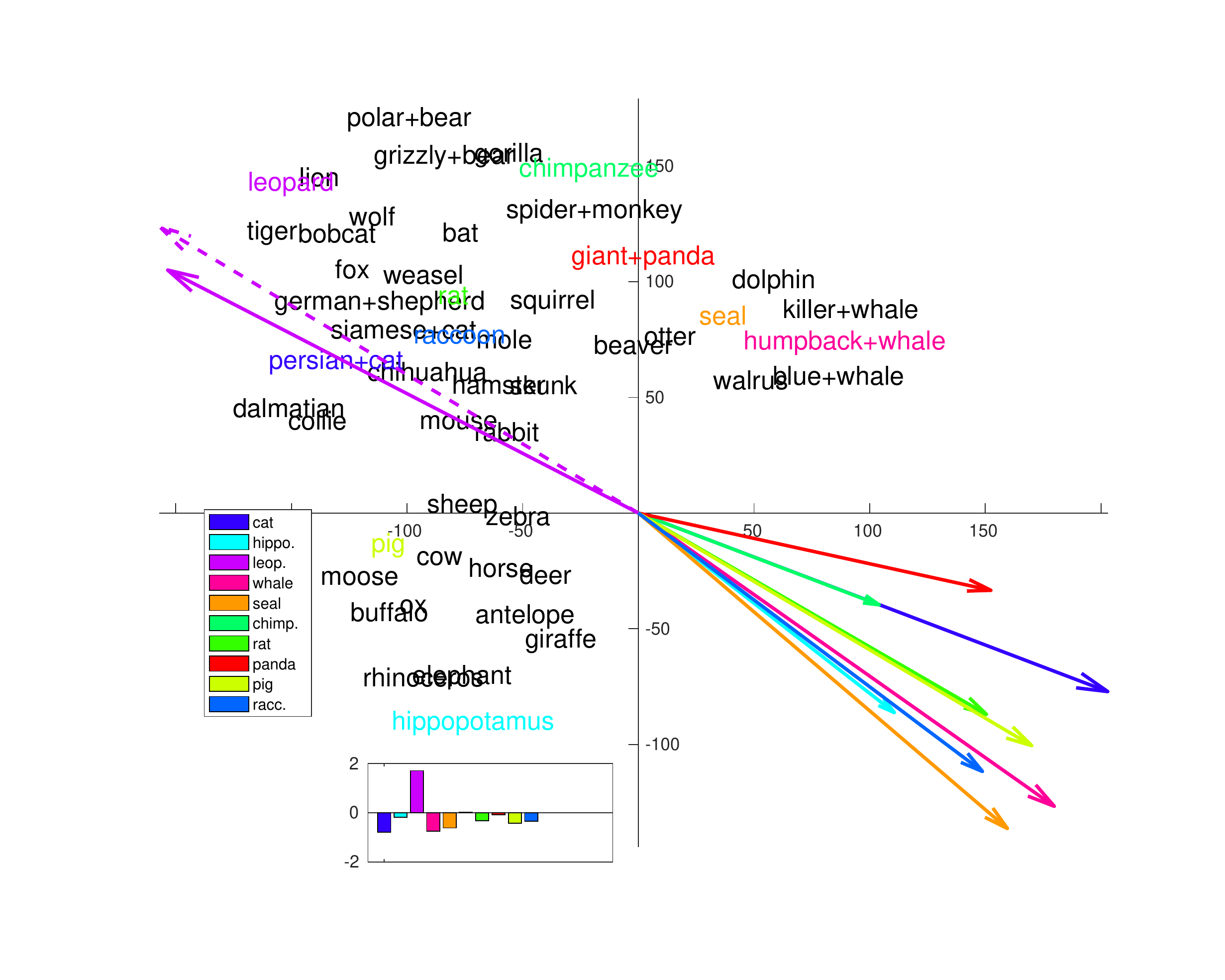}\label{fig:f1}}
	\hspace{15mm}
  \subfloat[Input image from unseen class Persian cat]{\includegraphics[width=0.4\textwidth,trim={2.8cm 2cm 1.8cm 1.6cm},clip]{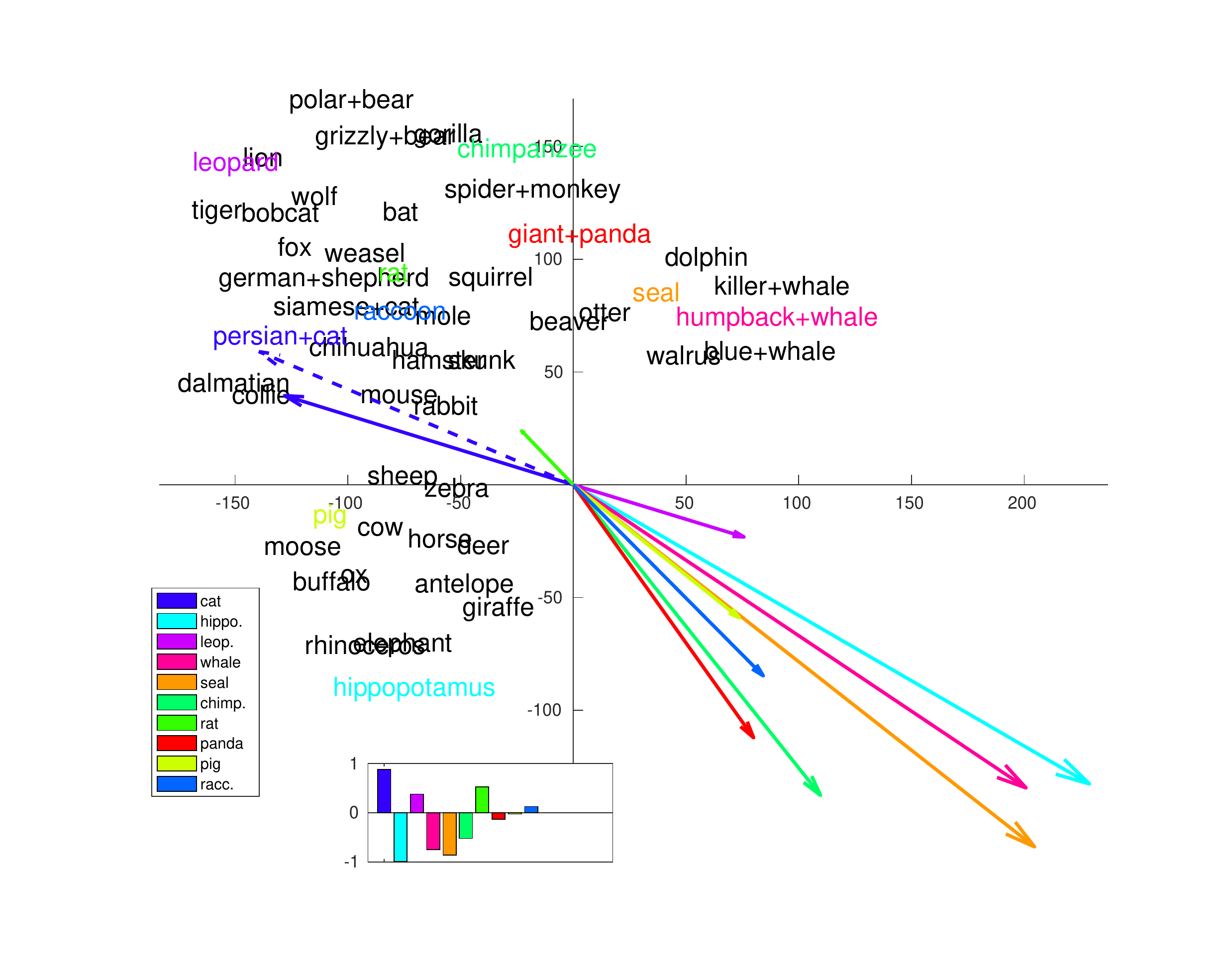}\label{fig:f2}}
  \caption{Visualization of class adapting principal directions (CAPD) on a 2D tSNE~\cite{tSNE_van2014} for illustration. Text labels on the plot represent the seen (black) and unseen (colored) semantic space embeddings in 2D space. CAPDs of the unseen classes are drawn with the same color of the unseen class label text. The bars indicate the responses of a semantic space embeddings projected onto their corresponding CAPDs. Our approach classifies an input to the class that has the maximum response. We also introduce an improved approach to use a reduced set of CAPDs (shown as dashed line) while obtaining better alignment with the correct unseen class embedding (see Sec. \ref{subsec:sparse}). (a) Leopard and (b) Persian cat.}
  \label{fig:pd}
\end{figure*}

Towards an ultimate visual object classification, this paper addresses three inherent handicaps of supervised learning approaches. The \textbf{first} one is the dependence on the availability of labeled training data. When object categories grow in number, sufficient annotations cannot be guaranteed for all objects beyond simpler and frequent single-noun classes. For composite and exotic concepts (such as American crow and auto racing paddock) not only the available images do not suffice as the number of combinations would be unbounded, but often the annotations can be made only by experts \cite{Krause_CVPR_2015,CUB_2011}. The \textbf{second} challenge is the appearance of new classes after the learning stage. In real world situations, we often need to deal with an ever-growing set of classes without representative images. Conventional approaches, in general, cannot tackle such recognition tasks in the wild. The \textbf{last} shortcoming is that supervised learning, in its customarily contrived forms, disregards the notion of wisdom. This can be exposed in the fact that we can identify a new object by just having a description of it, possibly leveraging its similarities with the previously learned concepts, without requiring an image of the new object \cite{Lampert_PAMI_2014}.  

In the absence of object annotations, zero-shot learning (ZSL) aims at recognizing object classes not seen at the training stage. In other words, ZSL intends to bridge the gap between the seen and unseen classes using semantic (and syntactic) information, which is often derived from textual descriptions such as word embeddings and attributes. Emerging work in ZSL attempt to predict and incorporate semantic embeddings to recognize unseen classes \cite{Hinton_NIPS_2009,Wang_CVPR_2013,Lampert_PAMI_2014,Yu_CVPR_2013,Mensink_CVPR_2014}. As noted in \cite{Jayaraman_NIPS_2014}, semantic embedding itself might be noisy. Instead of a direct embedding, some methods \cite{Akata_CVPR_2015,Xian_2016_CVPR,Qiao_2016_CVPR,Zhang_2016_CVPR} utilize global compatibility functions, e.g. a single projection in \cite{Zhang_2016_CVPR}, that project image features to the corresponding semantic representations. Intuitively, different seen classes contribute differently to describe each unseen class. Enforcing all seen and unseen classes into a single global projection undermines the subtle yet important differences among the seen classes. It eventually limits ZSL approaches by over-fitting to a specific dataset, visual and semantic features (supervised or unsupervised). Besides, incremental learning with newly added unseen classes using a global projection is also problematic due to its less flexibility. 

Traditionally, ZSL approaches (e.g., \cite{Changpinyo_2016_CVPR,Zhang_2015_ICCV,romera_ICML_2015}) assume that only the unseen classes are present in the test set. This is not a realistic setting for recognition in the wild where both unseen, as well as seen classes, can appear during the test phase. Recently \cite{Xian_CVPR_2017,Chao_ECCV_2016} tested several ZSL methods in generalized zero-shot learning (GZSL) settings and reported their poor performance in this real world scenario. The main reason of such failure is the strong bias of existing approaches towards seen classes where almost all test unseen instances are categorized as one of the seen classes. Another obvious extension of ZSL is few/one-shot learning (F/OSL) where few labeled instances of each unseen class are revealed during training. The existing ZSL approaches, however, do not scale well to the GZSL and FSL settings \cite{Akata_2016_CVPR,Changpinyo_2016_CVPR,Xian_2016_CVPR,Zhang_2016_CVPR,bucher_ECCV_2016}. 

To provide a comprehensive and flexible solution to ZSL, GZSL and FSL problem settings, we introduce the concept of principal directions that adapt to classes. In simple terms, CAPD is an embedding of the input image into the semantic space such that, when projected onto CAPDs, the semantic space embedding of the true class gives the highest response. A visualization of the CAPD concept is presented in Fig.~\ref{fig:pd}. As illustrated, the CAPDs of a Leopard (Fig.~\ref{fig:f1}) and a Persian cat image (Fig.~\ref{fig:f2}) point to their true semantic label embedding shown in violet and blue respectively, which gives the highest projection response in each case.


Our proposed approach utilizes three main sources of knowledge to generalize learning from seen to unseen classes. \textbf{First}, we model the relationships between the visual features and semantics for seen classes using the proposed `Class Adapting Principal Directions' (CAPDs). CAPDs are computed using class-specific discriminative models which are learned for each seen category in the `\emph{visual domain}' (Sec.~\ref{CAPD1}). \textbf{Second}, our approach effectively models the relationships between the seen and unseen classes in the `\emph{semantic space}' defined by CAPDs. To this end, we introduce a mixing transformation, which learns the optimal combination of seen semantics which are sufficient to reconstruct the semantic embedding of an unseen class (Sec.~\ref{CAPD2}). \textbf{Third}, we learn a distance metric for the seen CAPDs such that samples belonging to the same class are clustered together, while different classes are mapped further apart (Sec.~\ref{CAPD2}). This learned metric transfers cross domain knowledge from visual domain to semantic embedding space. Such a mapping is necessary because the class semantics, especially those collected from unsupervised sources (e.g., word2vec), can be noisy and highly confusing. The distance metric is then used to robustly estimate the seen-unseen semantic relationships.

\begin{SCfigure*}
   \includegraphics[width=1.3\linewidth,trim={1.3cm 15.5cm 5.4cm 4.3cm},clip]		{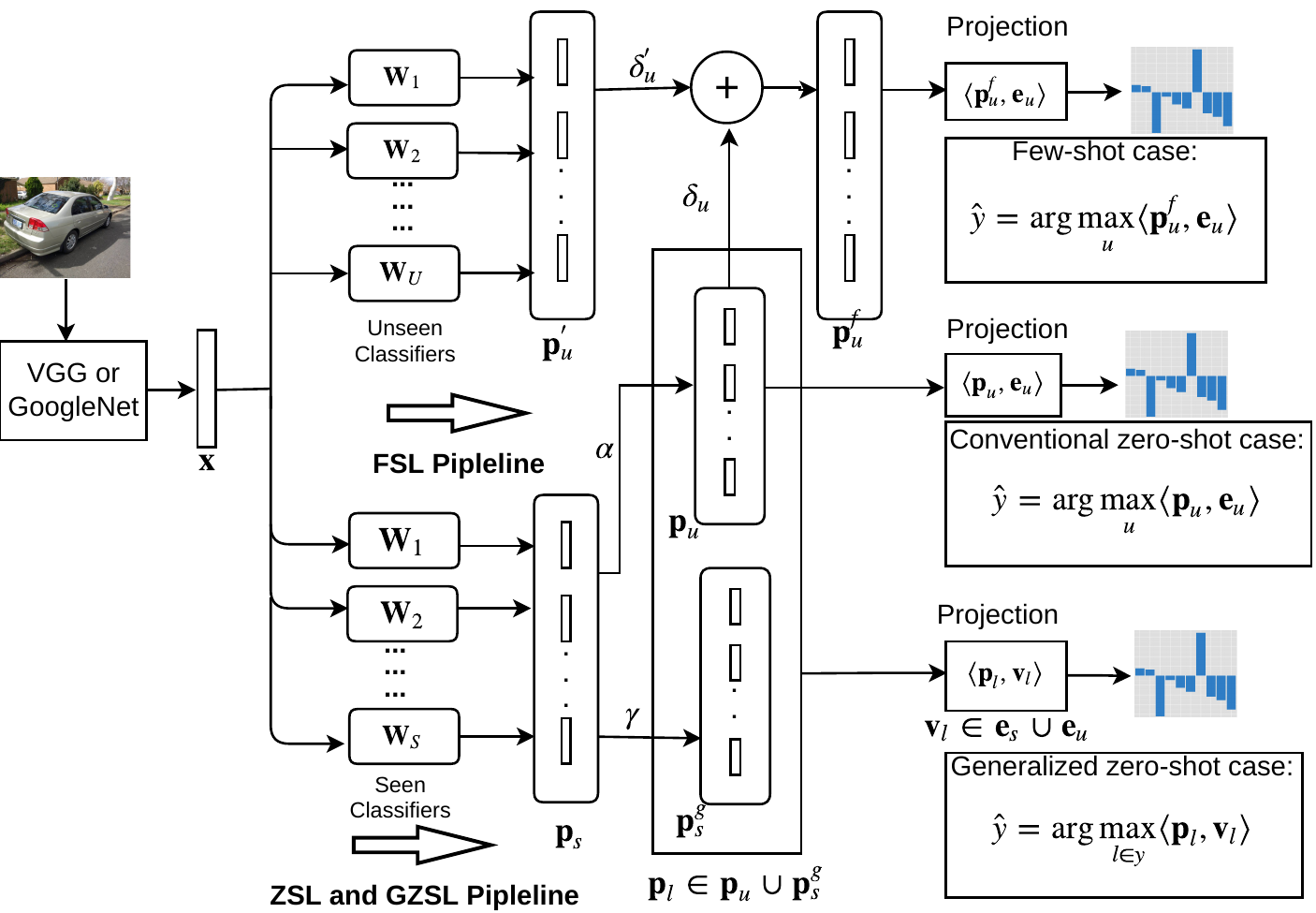}
   \caption{\small \protect\rule{0ex}{5ex}Overall pipeline of conventional ZSL, FSL and GZSL method. \textbf{Conventional ZSL:} An image is passed through a deep network to get an image feature $\x$. Then, $\x$ is fed to seen classifiers $\W_s$ to produce seen CAPDs, $\p_s$. Afterwards, unseen CAPDs, $\p_u$ are computed by linearly combining seen CAPDs using $\alpha$ (or $\beta$ for reduced case). Finally, prediction is done by computing the maximum projection response of $\p_u$ and unseen semantic embeddings $\mathbf{e}_u$. \textbf{FSL:} $\x$ is fed to unseen classifiers $\W_u$ to produce another version of unseen CAPDs $\p'_u$ which are combined with previously computed $\p_u$ through $\delta'_u$ and $\delta_u$ to find an updated version of unseen CAPDs $\p^f_u$. Final prediction is done by maximum response of $\p^f_u$ and $\mathbf{e}_u$.  \textbf{GZSL:} Seen CAPDs, $\p_s$ of conventional ZSL setting are generalized using $\gamma$ to produce generalized seen CAPDs, $\p^g_s$. For prediction, both $\p^g_s$ and $\p_u$ are considered for calculating maximum response of CAPDs and their corresponding semantic embeddings, $\mathbf{e}_s$ and $\mathbf{e}_u$.}
\label{fig:flow}
\end{SCfigure*}

While most of the approaches in the literature focus on specific sub-problems and do not generalize well to other related settings, we present a unified solution which can easily adapt to ZSL, GZSL and F/OSL settings. We attribute this strength to two key features in our approach: \textbf{a)}  a highly `modular learning' scheme and \textbf{b)} the two-way inter-domain `knowledge sharing'. Specifically for the GZSL,  we present a novel method to generalize seen CAPDs that avoids the inherent bias of prediction towards seen classes (Sec.~\ref{subsec:GZSL}). The generalized \emph{seen} CAPD balances the seen-unseen diversity in the semantic space, without any direct supervision from the visual data. In contrast to ZSL and GZSL, the F/OSL setting allows few or a single training instance of the unseen classes. This information is used to update \emph{unseen} CAPDs based on the learned relationships between visual and semantic domains for unseen classes (Sec.~\ref{subsec:FSL}). The overall pipeline of our learning and prediction process is illustrated in Fig. \ref{fig:flow}. 

We hypothesize that not all seen classes are instrumental in describing a novel unseen category. To validate this claim, we introduce a new constraint during the reconstruction of semantic embedding of the unseen classes. We show that automatically {reducing} the number of seen classes in the mixing process to obtain CAPD of each unseen class results in a significant performance boost (Sec.~\ref{subsec:sparse}). 
We perform extensive experimental evaluations on four benchmark datasets and compare with several state-of-the-art methods. Our results demonstrate that the proposed CAPD based approach provides superior performance in supervised and unsupervised settings of ZSL, GZSL and F/OSL.

To summarize, our main contributions are:
\begin{itemize}
\item We present a unified solution by introducing the notion of class adapting principal directions that enable efficient and discriminative embeddings of unseen class images in the semantic space.  
\item We propose a semantic transformation to link the embeddings for seen and unseen classes based on a learned distance measure. 
\item We provide an automatic solution to select a reduced set of relevant seen classes resulting in a better performance.
\item Our approach can automatically adapt to generalized zero-shot setting by generalizing seen CAPDs to match seen-unseen diversity.
\item Our approach is easily scalable to few/one-shot setting by updating the unseen CAPDs with newly available data.

\end{itemize}

\section{Related Work}

\textbf{Class Label Description:} It is a common practice to employ class label descriptions to transfer knowledge from seen to unseen class in ZSL. Such descriptions may come from either supervised or unsupervised learning settings. For the supervised case, class attributes can be one source as well \cite{aPY_2009,AwA_2009,SUN_2014,CUB_2011}. These attributes are often generated manually, which is a laborious task. As a workaround, word semantic space embeddings derived from a large corpus of unannotated text (e.g. from Wikipedia) can be used. Among such unsupervised word semantic embeddings, word2vec \cite{Mikolov_NIPS_2013,Mikolov_arXiv_2013} and GloVe \cite{Jeffrey_Glove_2014} vectors are frequently employed in ZSL \cite{Mubarak_2016_CVPR,Xian_2016_CVPR}. These ZSL methods are sometimes (arguably confusingly) referred as unsupervised zero-shot learning \cite{Al-Halah_2016_CVPR,Akata_2016_CVPR}. Supervised features tend to provide better performance than the unsupervised ones. Nevertheless, unsupervised features provide more scalability and flexibility since they do not require expert annotation. Recent approaches attempt to advance unsupervised ZSL by mapping textual representations (e.g. word2vec or GloVe) as attribute vectors using heuristic measures \cite{Kodirov_2015_ICCV,Al-Halah_2016_CVPR}. In our work, we use both types of features and evaluate on both supervised and unsupervised ZSL to demonstrate the strength of our approach.
    
\textbf{Embedding Space:} ZSL strategies aim to map between two different sources of information and two spaces: image and label embeddings. Based on the mapping scheme, ZSL approaches can be grouped into two categories. The \textbf{first} category is attribute/word vector prediction. Given an image, they attempt to approximate label embedding and then classify an unseen class image based on the similarity of predicted vector with unseen attribute/word vector. For example, in an early seminal work, \cite{Hinton_NIPS_2009} introduced a semantic output code classifier by using a knowledge base of attributes to predict unseen classes. \cite{Wang_CVPR_2013,Lampert_PAMI_2014} proposed direct and indirect attribute prediction methods via a probabilistic realization. \cite{Yu_CVPR_2013} formulated a discriminative model of category level attributes. \cite{Mensink_CVPR_2014} proposed an approach of transferring semantic knowledge from seen to unseen classes by a linear combination of classifiers. The main problem with such direct attribute prediction is the poor performance when noisy or biased attribute annotations are available. Jayaraman and Grauman \cite{Jayaraman_NIPS_2014} addressed this issue and proposed a discriminative model for ZSL. 

Instead of predicting word vectors, the \textbf{second} category of approaches learn a compatibility function between image and label embeddings, which returns a compatibility score. An unseen instance is then assigned to the class that gives the maximum score. For example, \cite{Akata_CVPR_2013} proposed a label embedding function that ranks correct class of unseen image higher than incorrect classes. In \cite{romera_ICML_2015}, authors use the same principle but an improved loss function and regularizer. Qiao \textit{et al.} \cite{Qiao_2016_CVPR} further improved the former approach by incorporating a component for noise suppression. In a similar work, Xian \textit{et al.} \cite{Xian_2016_CVPR} added latent variables in the compatibility function which can learn a collection of label embeddings and select the correct embedding for prediction. Our method also has similar compatibility function based on inner product of CAPD and corresponding semantic vector. The use of CAPDs provide an effective avenue to recognition.

\textbf{Similarity Matching:} This type of approaches build linear or nonlinear classifiers for each seen class, and then relate those classifiers with unseen classes based on class-wise similarity measures \cite{Changpinyo_2016_CVPR,Elhoseiny_ICCV_2013,Gavves_ICCV_2015,Mensink_CVPR_2014,Rohrbach_CVPR_2011}. Our method finds similar relation but instead of classifiers, we relate CAPDs of seen and unseen classes. Moreover, we compute this relation on a learned metric of semantic embedding space which let us consider subtle discriminative details.

\textbf{Few/One-shot Learning:} FSL has a long history of investigation where few instances of some classes are used as labeled during training \cite{Salakhutdinov_PAMI_2013,Fei_PAMI_2006}. Although ZSL problem can easily be extended to FSL, established ZSL methods are not evaluated in FSL settings. A recent work \cite{ReViSE_CoRR_2017} reports FSL performance of only two ZSL methods e.g. \cite{Socher_NIPS_2013,DeViSE_NIPS_2013}. In another work, \cite{Changpinyo_2017_ICCV} presented FSL results on ImageNet. In this paper, we extend our approach to FSL settings and compare our method with the reported performance in \cite{ReViSE_CoRR_2017}.

\textbf{Generalized Zero-shot Learning:}
GZSL setting significantly increases the complexity of the problem by allowing both seen and unseen classes during testing phase \cite{Xian_CVPR_2017,Chao_ECCV_2016,Changpinyo_2017_ICCV}. This idea is related to open set recognition problem where methods consider to reject unseen objects in conjunction with recognizing known objects \cite{Bendale_CVPR_2016,Jain_ECCV_2014}. In open set case, methods consider all unseen objects as one outlier class. In contrast, GZSL represents unseen classes as individual separate categories. Very few of the ZSL methods reported results on GZSL setting \SR{\cite{Changpinyo_2017_ICCV,Li_2017_CVPR,Xu_Matrix_CVPR_2017}}. \cite{DeViSE_NIPS_2013} proposed a joint visual-semantic embedding model to facilitate the generalization of ZSL. \cite{Socher_NIPS_2013} offered a novelty detection mechanism which can detect whether the test image came from seen or unseen category. Chao \textit{et al.} \cite{Chao_ECCV_2016} proposed a calibration mechanism to balance seen-unseen prediction score which any ZSL algorithm can adopt at decision making stage and proposed an evaluation method called Area Under Seen-Unseen accuracy Curve (AUSUC). Later, several other works \cite{Changpinyo_2017_ICCV,Xu_Matrix_CVPR_2017} adopted this evaluation strategy. In another recent work, Xian \textit{et al.} \cite{Xian_CVPR_2017} reported benchmarking results for both ZSL and GZSL performance of several established methods published in the literature. In this paper, we describe extension of our ZSL approach to efficiently adapt with GZSL settings.


\section{Our Approach}

\textbf{Problem Formulation:} Suppose, the set of all class labels is $\y = \y^{\se} \cup \y^{\un}$ where $\y^{\se}=\{1, ... , \rs \}$ and $\y^{\un} = \{\rs+1, ... , {\rs+\ru} \}$ are the sets of seen and unseen class labels respectively, with no overlap i.e., $\y^{\se} \cap \y^{\un} = \emptyset$. Here,  $\rs$ and $\ru$ denote the total number of seen and unseen classes, respectively. For all classes in the seen and unseen class sets, we have associated semantic class embeddings (either attributes or word vectors) denoted by the sets $\mathbf{E}^{\se} = \{\e_s : s \in \y^{\se}\}$ and $\mathbf{E}^{\un} = \{\e_u : u \in \y^{\un}\}$ respectively, where $\e_s, \e_u \in \mathbb{R}^d$. For every seen ($s$) and unseen ($u$) class, we have a number of instances denoted by $n_s$ and $n_u$ respectively. The matrices $\X_s = [\x_s^1, ..., \x_s^{n_s}]$ for $s\in \y^{\se}$, and $\X_u = [\x_u^1, ..., \x_u^{n_u}]$ for $u\in \y^{\un}$ represent the image features for the seen class $s$ and the unseen class $u$, respectively, such that $ \x_{s}, \x_{u} \in \mathbb{R}^k$. Below, we define the three problem instances addressed in this paper: 
\begin{itemize}
\item \textbf{Zero Shot Learning (ZSL):} The image features of the unseen classes $\X_u$ are not available during the training stage. The goal is to assign an unseen class label $u \in \y^{\un}$ to a given unseen image using its feature vector $\x_u$.

\item \textbf{Generalized Zero Shot Learning (GZSL):} The image features of the unseen classes $\X_u$ are not available during the training stage similar to ZSL. The goal is to assign a class label $l \in \y$ to a given image using its feature vector $\x$. Notice that, the true class of $\x$ may belong to either a seen or an unseen class.

\item \textbf{Few/One Shot Learning (FSL):} Only a few/one randomly chosen image features from $\X_u$ are available as labeled examples during the training stage. The goal is same as the ZSL setting above.
\end{itemize}

In Secs.~\ref{subsec:CAPD_for_Classes} and \ref{subsec:sparse}, we first provide a general framework of our approach mainly focused on ZSL. Afterwards, in Secs.~\ref{subsec:FSL} and \ref{subsec:GZSL} we extend our approach to FSL and GZSL settings, respectively. Before describing our extensive experimental evaluations in Sec.~\ref{sec:Experiments}, we also provide an in-depth comparison with the existing literature in Sec.~\ref{subsec:comp_analysis}. 

\subsection{Class Adapting Principal Direction}
\label{subsec:CAPD_for_Classes}
We introduce the concept of `Class Adapting Principal Direction' (CAPD), which is a projection of image features onto the semantic space. The CAPD is computed for both  seen and unseen classes, however the derivation of the CAPD is different for both cases. In the following, we first introduce  our approach to learn CAPDs for seen classes and then use the learned principal directions to derive CAPDs for unseen classes.

\subsubsection{Learning CAPD for Seen Classes}\label{CAPD1}
For a given image feature $\x_s$ belonging to the seen class $s$, we define its CAPD $\p_s$ in terms of a linear mapping parametrized by $\W_s$ as,
\begin{equation}
    \p_s = \W_s^T \x_s.
\label{eq:pd_seen}
\end{equation}
Our goal is to learn the  class-specific weights $\W_s$ such that the output principal directions are highly discriminative in the semantic space (rather than the image feature space). To this end, we introduce a novel loss function which uses the corresponding semantic space embedding $\e_s$ of  seen class $s$ to achieve maximum separability.

\textbf{Proposed Objective Function:}
Given the training samples $\X_s$ for the seen class $s$, $\W_s$ is learned such that the projection of $\p_s$ on the semantic space embedding $\e_s$, defined by the inner product $\langle \p_s,\e_s \rangle$, generates a strong response. Precisely, the following objective function is minimized: 
\begin{equation}
      \min_{\W_s} \frac{1}{\kappa} \sum_{c=1}^{\rs} \sum_{m = 1}^{n_c} \log \Big( 1 + \exp \big\{ L(\x_c^m;\W_s) \big\} \Big) + \frac{\lambda_{s}}{2} \parallel \W_s \parallel_2^2
\label{eq:seenwights}
\end{equation}
where $L$ is the cost for a specific input $\x_c^m$, $\lambda_s$ is the regularization weight set using cross validation and $\kappa = \sum_{c=1}^{\rs}n_c$. We define the cost $L$ as:
\begin{equation*}
   L(\x^m_{c};\W_s)=\left\{
  \begin{array}{@{}ll@{}}
    \langle \p_s,\e_c \rangle - \langle \p_s,\e_s \rangle, & c \neq s \\
    \langle \p_s,\frac{1}{\rs-1}\sum\limits_{\substack{t\neq s}} \e_t \rangle - \langle \p_s,\e_s \rangle, &  c = s
  \end{array}\right.
\end{equation*}
In the above loss function, two different scenarios are tackled depending on whether the training samples (image features) are from the same (positive) or different (negative) classes. For the \textbf{negative} samples ($c\neq s$), the projection of $\p_s$ on the correct semantic embedding $\e_s$ is maximized while its projection on the incorrect semantic embedding $\s_c$ is minimized. For the \textbf{positive} samples ($c=s$), our proposed formulation directs the projection on the correct semantic embedding $\e_s$ to be higher than the average response of projections on the incorrect semantic embeddings. In both cases, $\langle \p_s,\e_s \rangle$ is constrained to produce a high response.  Our loss formulation is motivated by \cite{Mubarak_2016_CVPR}, with notable differences such as the class-wise optimization, explicit handling of positive samples and the extension of their ranking loss for image tagging to the ZSL problem. 

We optimize Eq.~\ref{eq:seenwights} by Stochastic Gradient Descent (SGD) to obtain $\W_s$ for each seen class. Note that, $\p_s=\W_s^T \x_c^m$ in the above cost function, thus for any sample $\x_c^m$, $\p_s$ changes when $\W_s$ is updated at each SGD iteration. Also, the learning process of $\W_s$ for each seen class is independent of other classes. Therefore, all $\W_s$ can be learned jointly in a parallel fashion.  Once the training process is complete, given an input visual feature $\x^m_c$, we generate one CAPD $\p_s$  for each seen class using Eq.~\ref{eq:pd_seen}. As a result, $\mathbf{P}^{\se} = [\p_1 ... \p_{\rs}] \in \mathbb{R}^{d\times \rs}$ accumulates the CAPDs of all the seen classes. Each CAPD is the mapped version of the image feature on the class specific semantic space. The CAPD vector and its corresponding semantic space embedding vector point to similar direction if the input feature belongs to the same class.

\subsubsection{Learning CAPD for Unseen Classes} \label{CAPD2}
In ZSL settings, the images of the unseen classes are not observed during the training. For this reason, we cannot directly learn a weight matrix to calculate $\p_u$ using the same approach as $\p_s$. Instead, for any unseen sample, we propose to approximate $\p_u$ using the seen CAPD of the same sample. Here, we consider a bilinear map, in particular, a linear combination of the seen class CAPDs to generate the CAPD of the unseen class $u$:
\begin{equation}
\p_u = \sum_{s=1}^{\rs} \theta_{s,u} \p_s = \mathbf{P}^{\se} \mathbf{\theta}_u
\label{eq:pdrelation}
\end{equation}
where, $\mathbf{\theta}_u = [\theta_{1,u} ... \theta_{\rs,u}]^T \in \mathbb{R}^{\rs}$ is the coefficient vector that, in a way, aggregates the knowledge of seen classes into the unseen one. 
The computation of $\mathbf{\theta}_u$ is subject to the relation between CAPDs and semantic embeddings of classes. We detail our approach to approximate $\mathbf{\theta}_u$ below.

\textbf{Metric Learning on CAPDs:}
The CAPDs reside in the semantic embedding space. In this space, we learn a distance metric to better model the similarities and dissimilarities among the CAPDs. To this end, we assemble the sets of similar $\mathbf{A}$ and dissimilar $\mathbf{\bar{A}}$ pairs of CAPDs that correspond to the pairs of training samples belonging to the same and different seen classes, respectively. Our goal is to learn a distance metric $d_{\mathbf{M}}$ such that the similar CAPDs are clustered together and the dissimilar ones are mapped further apart. We minimize the following objective which maximizes the squared distances between the  minimally separated dissimilar pairs: 
\begin{align} 
      \max_{\mathbf{M}} \min_{(i,j)\in \mathbf{\bar{A}}} d^2_{\mathbf{M}} (\p_i,\p_j)   \quad 
s.t. \sum_{(i,j)\in \mathbf{A}} d^2_{\mathbf{M}} (\p_i,\p_j) \leq 1 
\label{eq:metriclearning}
\end{align}
where $d_{\mathbf{M}} = \sqrt{(\p_i-\p_j)^T\mathbf{M}(\p_i-\p_j)}$ is the Mahalanobis distance metric \cite{Ying_JMLR_2012}.  After training, the most confusing dissimilar CAPD pairs are pulled apart while the similar CAPDs are clustered together by learning an optimal distance matrix $\mathbf{M}$.

Our intuition is that, given a learning metric $\mathbf{M}$ in the semantic embedding space, the relation between the semantic label embeddings of the seen $\e_s$ and the unseen classes $\e_u$ is analogous to that of their principal directions. Since the semantic label embedding of unseen classes are available, we can estimate their relation with the seen classes. For simplicity, we consider a linear combination of semantic space embeddings:
\begin{equation}
\hat{\e}_{u} = \sum_{s=1}^{\rs} \alpha_{s,u} \e_s = \mathbf{E^{\se}\alpha}_u
\label{eq:uvrelation}
\end{equation}
where, $\hat{\e}_{u}$ is the approximated semantic embedding of $\e_u$ corresponding to unseen class $u$. We compute $\mathbf{\alpha}_u = [\alpha_{1,u} ... \alpha_{\rs,u}]^T \in \mathbb{R}^{\rs}$ by solving:
\begin{equation}
 \min_{\alpha_u} \quad (\hat{\e}_{u}- \e_u)^T \mathbf{M} (\hat{\e}_{u} - \e_u) + \frac{\lambda_u}{2}  \parallel\alpha_u\parallel_2^2
\label{eq:seenwordalpha}
\end{equation}
where $\lambda_u$ is a regularization parameter which is set via cross validation.

As we mentioned above, using the learned metric $\mathbf{M}$, the relationship between the seen-unseen semantic embeddings $\alpha_u$ is analogous to the relationship between the seen-unseen CAPDs $\theta_u$, thus $\theta_u \approx \alpha_u$. Accordingly, we approximate the unseen CAPDs with seen CAPDs by rewriting Eq.~\ref{eq:pdrelation} as:
\begin{equation}
	\p_u \approx \mathbf{P}^{\se}\alpha_u.
    \label{eq:pdrelation2}
\end{equation}
We derive a CAPD, $\p_u$ for each unseen class using Eq.~\ref{eq:pdrelation2}. In test stage of ZSL setting, we assign a given image feature $\x$ to an unseen class using the maximum projection response:
\begin{equation}
\label{eq:unseen_prediction}
\hat{y} = \arg \max_u \langle \p_u, \e_u \rangle
\end{equation}

\begin{figure}[t]
  \begin{center}
    \includegraphics[width=.9\linewidth]{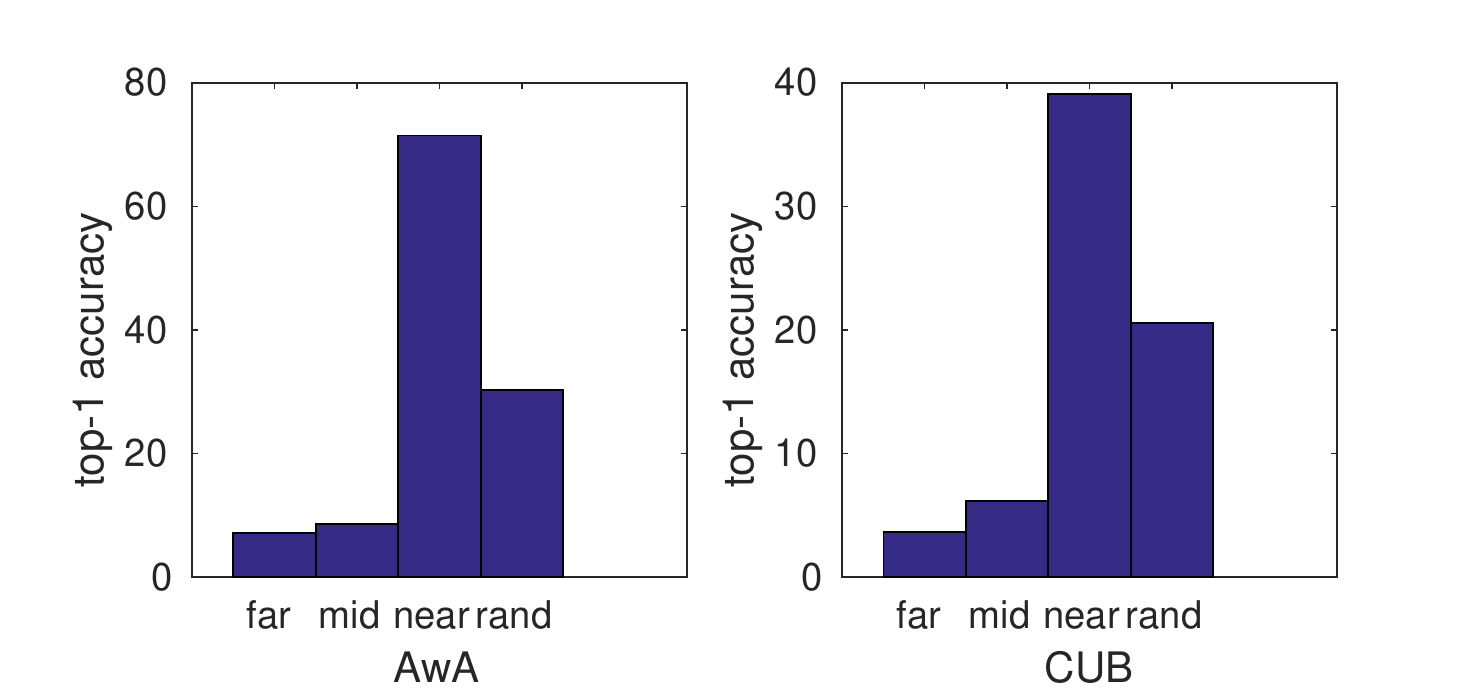}
  \end{center}
  \vspace{-1.6em}
   \caption{Experiments with the farthest away, mid-range, nearest, and randomly chosen seen classes, using one third of the total seen classes in each case. Image features are obtained using VGG-verydeep-19 and semantic space vectors are derived from attributes. As shown, the semantic space embeddings of the seen classes that are {\bf near} to the embedding of the unseen class provide more discriminative representations.
   }
\label{fig:closefar}
\end{figure}

\subsection{Reduced Set Description of Unseen Classes}
\label{subsec:sparse}
When describing a novel object, we often resort to relating it with the similar known object categories. It is intuitive that a subset of the known objects is sufficient for describing an unknown one. 

We incorporate this observation by proposing a modified version of  Eq.~\ref{eq:uvrelation}. The term $\mathbf{\alpha}_u$ contains the contribution of each seen class to describe the unseen class $u \in \y^{\un}$ by reconstructing $\e_u$ using all seen classes semantic label embeddings. We reconstruct $\e_u$ by only a small number of seen classes ($\rn < \rs$). These $\rn$ seen classes can be selected using any similarity measure (Mahalanobis distance in our case). The reconstruction of $\e_u$ becomes:
\begin{equation}
\hat{\e}_{u} = \sum_{i=1}^{\rn} \beta_{i,u} \e_i
\label{eq:uvrelation2}
\end{equation}
Here, $\mathbf{\beta}_u\in \mathbb{R}^{\rn}$ is the coefficients of selected seen classes. We learn $\mathbf{\beta}_u$ by a similar minimization objective as in the Eq.~\ref{eq:seenwordalpha}. By replacing $\alpha_u$ with $\beta_u$ in the Eq.~\ref{eq:pdrelation2}, it is possible to compute the CAPD of unseen class $u$ using a reduced set of seen classes. Such CAPDs are shown in Fig.~\ref{fig:pd} in dashed lines.

\textbf{Appropriate Choice of Seen Classes:} 
In Fig.~\ref{fig:closefar}, we show comparisons when different approaches are used to select a subset of seen classes to describe the unseen ones. The results illustrate that the seen classes having the semantic space embeddings close to that of a particular unseen class are more suitable to describe it. Here, we considered $\rn$  nearest neighbors of the unseen class semantic vector $\e_u$ using the Mahalanobis distance. Using a less number of seen classes is inspired by the work Norouzi \emph{et al.} \cite{norouzi_arXiv_2013} where they applied convex combination of selected semantic embedding vector based on outputs of the softmax classifier of corresponding seen classes. The main drawback of their approach is that the softmax classifier output does not take the semantic embeddings into consideration, which can ignore important features when describing the unseen class. Instead, our method performs an independent optimization (Eq. \ref{eq:seenwordalpha}) that jointly considers image feature, CAPD and semantic embedding relations via the learned metric $\mathbf{M}$. As a result, the proposed strategy is better able to determine the optimal combination of selected seen semantic embeddings (see Sec.~\ref{comp_zsl}).

\textbf{Automatic $\rn$ Selection for Each Unseen Class:} While \cite{norouzi_arXiv_2013} proposed a \textit{fixed} number of selected seen classes to describe an unseen class, we suggest a novel technique to automatically select the number of most informative seen classes ($\rn$). 

\begin{figure}[t]
  \begin{center}
  \includegraphics[width=.8\linewidth,trim={.1cm .1cm 1cm .4cm},clip]{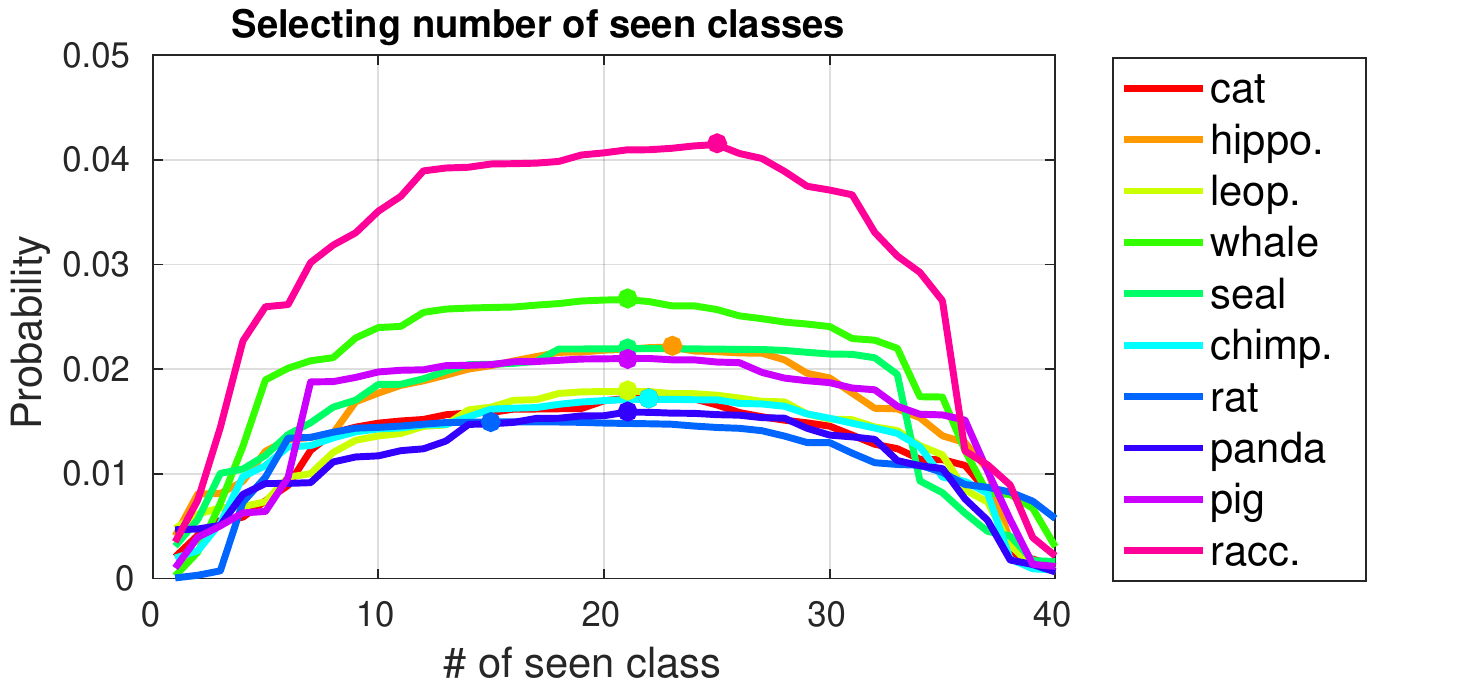}
  \end{center}
  \vspace{-1.2em}
   \caption{PDF of distances using the normal distribution with zero mean and a unit standard deviation for each unseen class. (GoogLeNet features and the word2vec semantic embedding for AwA dataset)}
\label{fig:autoseenselect}
\end{figure}

\emph{First}, for an unseen class semantic embedding $\e_u$, we calculate the Mahalanobis distances (using learned metric $\mathbf{M}$) from $\e_u$ to all $\e_s$ and perform mean normalization. Then, we apply kernel density estimation to obtain a probability density function (pdf) for the normalized distances. Fig.~\ref{fig:autoseenselect} shows the pdf for each unseen semantic embedding vector of the AwA dataset. For a specific unseen class, the number of seen classes with the highest probability score is assigned as the value of $\rn$. Unlike \cite{norouzi_arXiv_2013}, this scheme allows choosing a \textit{variable} number of the seen classes for different unseen classes. In Sec.~\ref{sec:reducedset} of this paper, we have reported an estimation of the average numbers of seen classes selected for the tested unseen classes.

\textbf{Sparsity:} Using a reduced number of the seen classes in Eq.~\ref{eq:uvrelation2} indirectly imposes sparsity in the coefficient vector $\mathbb{\alpha}_u$ in the Eq.~\ref{eq:uvrelation}. This is similar to Lasso $(\ell_1)$ regularization (instead of $\ell_2$ regularization) in the loss function in Eq.~\ref{eq:seenwordalpha}. We observe that the above selection solution is more efficient  and accurate than the Lasso-based regularization. This is because the proposed solution is based on the intuition that the semantic embedding of an unseen class can be described by closely embedded seen classes. In contrast, Lasso is a general approach and do not consider any domain specific semantic knowledge.

Having discussed the ZSL setting in Secs.~\ref{subsec:CAPD_for_Classes} and \ref{subsec:sparse} above, we present the extension of CAPDs to the GZSL problem.

\subsection{Generalized Zero-shot Learning}
\label{subsec:GZSL}
ZSL setting considers only unseen class images during the test phase. This setting is less realistic, because new images can belong to both seen and unseen classes. To address this scenario, generalized ZSL (GZSL) has recently been introduced as a new line of investigation \cite{Xian_CVPR_2017,Chao_ECCV_2016}. Recent works suggest that most of the existing ZSL approaches fail to cope up with the GZSL setting. When both seen and unseen classes come into consideration for prediction, the prediction score function becomes highly biased towards seen classes because only seen classes were used for training. As a result, majority of the unseen test instances are misclassified as seen examples. In other words, this bias notably decreases the classification accuracy on unseen classes while maintains relatively high accuracy on seen classes. To solve this problem, available techniques attempts to estimate the prior probability of an input belonging to either a seen or an unseen class \cite{Socher_NIPS_2013,Chao_ECCV_2016}. However, this scheme heavily depends on the original data distribution used for training. 

Considering the above aspects, a competent GZSL method should possess the following properties:
\begin{itemize}
\item \textit{Equilibrium:} It should be able to balance seen-unseen diversity so that the performances of both seen and unseen classes achieve a balance.
\item \textit{Reduced data dependency:} It should not receive any supervision signal (obtained from either training or validation set images) determining the likelihood of an input belonging to seen or unseen class.
\item \textit{Consistency:} It should retain its performance on the conventional ZSL setting as well.
\end{itemize}
In this work, we propose a novel GZSL algorithm to adequately address these challenges. 

\textbf{Generalized CAPD for Seen Class:}
In Sec. \ref{subsec:CAPD_for_Classes}, we described the CAPD of seen classes for a given input image is $\mathbf{P}^{\se} = [\p_1 ... \p_{\rs}]$. Each seen CAPDs is obtained using the class-wise learned classifier matrix $\W_s$. It is obvious that each $\W_s$ is biased to seen class `$s$'. For the same reason, each $\p_s$ is also biased to class `$s$'. Since there was no seen instance available during the testing phase in conventional ZSL setting, seen CAPDs were not used for prediction (Eq. \ref{eq:unseen_prediction}). Therefore, the inherent bias of seen CAPDs was not affecting ZSL performance. In contrast, for GZSL settings, all seen and unseen CAPDs are considered for prediction. Thus, biased seen CAPDs will dominate as expected and significantly affect the unseen class performances. To solve this problem, we propose to develop a generalized version of each seen CAPD as follows:
\begin{equation}
\centering
\mathbf{p}^{g}_s= \mathbf{P}^{\se} \gamma_s,
\end{equation}
where,  $\gamma_s$ denotes a parameter vector for seen class `$s$'.

\textbf{Proposed Objective Function:} Our hypothesis is that the bias towards seen classes that causes high scores during prediction can be resolved using the semantic information of classes. To elaborate, $\gamma_s$ is computed solely in semantic label embedding domain and later applied to generalize CAPD of seen class instances. We minimize the squared difference of two complementary losses to obtain $\mathbf{\gamma} = [\gamma_{1} ... \gamma_{\rs}] \in \mathbb{R}^{\rs \times \rs}$, as:
\begin{multline}
\min_{\gamma} \parallel\overbrace{\frac{1}{\rs}\sum_{s=1}^{\rs} (\mathbf{E}^{\se}\gamma_s-\e_s)^2}^\text{mean generalized seen loss}  -  \overbrace{\frac{1}{\ru}\sum_{u=1}^{\ru}(\mathbf{E}^{\se}\mathbf{\alpha}_u-\e_u)^2}^\text{mean unseen reconst. loss}\parallel_2^2  \\ 
+ \frac{\lambda_\gamma}{2} \sum_{s=1}^{\rs} \parallel\gamma_s\parallel_2^2,
\label{eq:gzsl_loss}
\end{multline}
where $\lambda_{\gamma}$ is the regularization weight set using cross validation.

The objective function in  Eq.~\ref{eq:gzsl_loss} minimizes the squared difference between the mean of two loss components. The \textbf{first} component is the mean generalized seen loss which measures the reconstruction accuracy of seen class embedding $\e_s$ using the generalization parameters $\gamma_s$. The \textbf{second} component measures the reconstruction accuracy of unseen class embedding $\e_u$ from seen classes. 
By reducing the squared difference between these two components, we indirectly balance the distribution of seen-unseen diversity which effectively prevents the domination of seen classes in the GZSL setting (the `\emph{equilibrium}' property). The interesting fact is that our proposed generalization mechanism does not directly use CAPDs, yet it is strong enough to stabilize the CAPD of different classes during the prediction stage (the `\emph{less data dependence}' property). Furthermore, the formulation does not affect the computation of unseen CAPDs i.e. $\p_u$ which preserves the conventional ZSL performance (the `\emph{consistency}' property).

\textbf{Prediction:} For a given image feature $\x$, we can derive generalized CAPDs of seen classes $\p^{g}_s$ and CAPD of unseen classes $\p_u$ using the description in Sec. \ref{subsec:sparse}. In test stage, we consider both the projection responses of seen and unseen classes to predict a class.
$$\hat{y} = \arg \max_{l \in \y} \langle \p_l, \mathbf{v}_l \rangle$$ where, $\p_l \in \p_u \cup \p_s^{g}$ and $\mathbf{v}_l \in \e_s \cup \e_u$.

\subsection{Few-shot Learning}
\label{subsec:FSL}

The few-shot learning (FSL) is a natural extension of ZSL. While ZSL considers no instance of an unseen class during training, FSL relaxes this restriction by allowing a few instances of an unseen class as labeled during the training process. Another variant of FSL is called one-shot learning (OSL), which allows exactly one instance of an unseen class (instead of few) as labeled during training. An ideal ZSL approach should be able to benefit from the labeled data for unseen classes under F/OSL settings. In this section, we explain how our approach is easily adaptable to FSL.

\textbf{Updated CAPD for Unseen Class.} In ZSL setting, for a given input image feature, we can calculate the unseen CAPD, $\p_u$ for every  unseen class `$u$'. Now, in the FSL setting, we optimally use the newly available labeled unseen data to update $\p_u$. To this end, new classifiers $\W_u$ are learned for each  unseen class `$u$' similar to the case of seen classes (Sec.~\ref{subsec:CAPD_for_Classes}).
For a given image feature, $\x$, we can calculate unseen CAPDs by $\p'_u = \W^T_u \x$. These CAPDs are fused with $\p_u$, which were derived from the linear combination of seen CAPDs (Eq.~\ref{eq:pdrelation2}). The updated CAPD for   unseen class `$u$' is represented as $\p^{f}_u$, given by:
\begin{equation}
\p^f_u = \delta_{u}\p_u + \delta'_{u}\p'_u, \qquad s.t. \;\; \delta_{u} + \delta'_{u} = 1
\label{eq:fewshotcombination}
\end{equation}
where, $\delta_u$ and $\delta'_{u}$ are the contribution of the respective CAPDs to form an updated CAPD of an unseen class. During prediction, we use $\p^f_u$ instead of $\p_u$ in Eq. \ref{eq:unseen_prediction}.

\textbf{Calculation of $\delta_{u}$ and $\delta'_{u}$:} The weights $\delta_u$ and $\delta'_{u}$ are set using training data such that they encode the reliability of $\p_u$ and $\p'_u$ respectively.
Recall that our prediction is based on the strength of projection of a CAPD on the semantic embedding vector. Therefore, we need to maximize the correspondence between a CAPD and the correct semantic embedding vector i.e., a high $\langle \p_u, \e_u \rangle$.
The unseen CAPD among $\p_u$ and $\p'_u$ that provides higher projection response with $u^{th}$ unseen class semantic vector gets a strong weight during the combination in Eq. \ref{eq:fewshotcombination}.

We derive $\p_u$ and $\p'_u$ for each training image feature, $\x \in \X^{\se} = \{\X_s : s \in \y^{\se}\}$, and the classification matrix of unseen class `$u$'. Then, we find the summation of maximum projection response of the CAPD (either $\p_u$ or $\p'_u$) with its respective semantic vector. This maximum projection response finds the response of most similar (or confusing) unseen class of any image. The summation of this response across all training images can estimate the overall quality of CAPDs from the two sources. Finally, we normalize the summations to get $\delta_{u}$ and $\delta'_{u}$ as follows:
$$
\delta_{u} = \frac {\sum_{\x \in \X^{\se}} \max_u \langle\p_u, \e_u\rangle} {\sum_{\x \in \X^{\se}} \max_u \langle\p_u,\e_u\rangle + \sum_{\x \in \X^{\se}} \max_u \langle\p'_u,\e_u\rangle},
$$
$$
\delta'_{u} = \frac {\sum_{\x \in \X^{\se}} \max_u \langle\p'_u,\e_u\rangle} {\sum_{\x \in \X^{\se}} \max_u \langle\p_u,\e_u\rangle + \sum_{\x \in \X^{\se}} \max_u \langle\p'_u,\e_u\rangle}.
$$


\section{Comparison with Related Work}\label{subsec:comp_analysis}
\subsection{ZSL Settings}
Our method has similarities with two streams of previous efforts on ZSL. Here, we discuss the significant differences. 

In terms of class-specific learning, a number of recent studies \cite{Changpinyo_2016_CVPR,norouzi_arXiv_2013} report competitive performances when they rely on handcrafted attributes (`\emph{supervised}' source). However, we observe that these methods fail  when they use `\emph{unsupervised}' source of semantics (e.g. word2vec and GloVe). The underlying reason is that they do not leverage on the semantic information during the training of the classifiers. Moreover, the attribute set is less noisy than unsupervised source of semantics. Although our work follows the same spirit, the main novelty lies in using the semantic embedding vectors explicitly during the learning phase for each individual class. This helps the classifiers to easily adapt themselves to a wide variety of semantic information sources, e.g. attributes, word2vec and GloVe. 

Another body of work \cite{Xian_2016_CVPR,Akata_CVPR_2015} considers semantic information during the training process. However, these approaches do not take the benefits of class-specific learning. Using a single classifier, they compute a global projection. Generalizing all classes by one projection is restrictive and it fails to encompass subtle variations among classes. These approaches do not leverage the flexibility of suppressing irrelevant seen classes while describing an unseen class. Besides, the semantic label embeddings are subject to tuning based on the visual image features. As they cannot learn any metric on semantic embedding space, these methods fail to work accurately across different semantic embeddings. In contrast, by taking the benefits of class-specific learning, our approach computes CAPD for each classifier that can significantly enhance the learned discriminative information. In addition, our approach describes the unseen class with automatically selected informative seen classes and learns a metric on the semantic embedding space to further fine-tune the semantic label information.

We also like to point out that the existing methods seem to overfit on a specific dataset, specific image features, and specific semantic features (supervised-attributes or unsupervised-GloVe). Our method, on the other hand, consistently provides improved performance across all the different problem settings. 

\subsection{GZSL settings}
We automatically balance the diversity of seen-unseen classes in an unsupervised way, without strongly relying on CAPD or image visual feature. Previous efforts used a supervision mechanism either from training or validation image data to determine if any input image belongs to a seen or an unseen class. Chao et al. \cite{Chao_ECCV_2016} proposed a calibration based approach to rescale the seen scores and evaluated using Area Under Seen-Unseen accuracy Curve (AUSUC) \cite{Changpinyo_2016_CVPR,Xu_Matrix_CVPR_2017}. As prediction scores of GZSL are strongly biased to seen classes, they proposed to calibrate seen scores by adding a constant negative bias termed as a calibration factor. This factor is calculated on a validation set and works as a prior likelihood of a data point being from a seen/unseen class. The drawback of such an approach is that it acts as a post-processing mechanism applied at the decision making stage, not dealing with the generalization at the basic algorithmic level.

Another alternative work, CMT method \cite{Socher_NIPS_2013} incorporates a novelty detection approach which estimates the outlier probability of an input image. Again, the outlier probability is determined using training images which provides an extra image-based supervision to GZSL model. In contrary, our method considers the seen-unseen biasness in the semantic space at the algorithmic level. 
The overall prediction scores are then balanced to remove the inherent biasness towards the seen classes. We show that such an approach can be useful for both supervised attributes and unsupervised word2vec/GloVe as semantic embedding information. As our approach does not follow the post-processing strategy like \cite{Chao_ECCV_2016, Changpinyo_2016_CVPR,Xu_Matrix_CVPR_2017}, we do not evaluate our work with AUSUC. In line with the recommendation in \cite{Xian_2016_CVPR}, we use harmonic mean based approach for GZSL evaluation.

\section{Experiments}
\label{sec:Experiments}
\textbf{Benchmark Datasets:} We use four standard datasets for our experiments; aPascal \& aYahoo (aPY) \cite{aPY_2009}, Animals with Attributes (AwA) \cite{AwA_2009}, SUN attributes (SUN) \cite{SUN_2014}, and Caltech-UCSD Birds (CUB) \cite{CUB_2011}. The statistics of these datasets are given in Table \ref{tab:dataset}. We follow the standard protocols (seen/unseen splits of classes) used in the literature. To be specific, we have exactly followed \cite{Xian_2016_CVPR} for AwA and CUB datasets, \cite{Zhang_2015_ICCV,Zhang_2016_CVPR} for aPY and SUN-10 and \cite{Changpinyo_2016_CVPR} for SUN. To increase the complexity of GZSL task for SUN, we used a different of seen/unseen split introduced in \cite{Changpinyo_2016_CVPR}. In line with the standard protocol, the test images correspond to only unseen classes in ZSL settings. In Few/One-shot settings, we randomly choose three/one instances per unseen class to use in training as labeled examples. Again, in GZSL settings, we perform a 80-20\% split of each seen class instances; 80\% portion is used in training and rest 20\% for testing in conjunction with all unseen test data. We report the average results of 10 random trails for Few/One shot or GZSL settings. In a recent work, Xian et al. \cite{Xian_CVPR_2017} proposed a different seen/unseen split for the above mentioned four datasets. We perform GZSL experiments on that setting as well.


\begin{table}[t]
  \begin{center}
    \begin{tabular}{|l|c|r|r|r|}
    \hline
    Dataset & seen/unseen & \# image  &\ \# train & \# test \\
    \hline\hline
    aPY\cite{aPY_2009} & 20/12 & 15,339  & 12,695 & 2,644\\
    AwA\cite{AwA_2009} & 40/10 & 30,475 & 24,518 & 6,180\\
    SUN-10\cite{SUN_2014} & 707/10 & 14,340 & 14,140 & 200\\
    SUN\cite{SUN_2014} & 645/72 & 14,340 & 12,900 & 1,440\\
    CUB\cite{CUB_2011} & 150/50 & 11,788 & 8,855 & 2,933\\
    \hline
    \end{tabular}
  \end{center}
  \vspace{-1em}
  \caption{Statistics of the benchmark datasets.}
  \label{tab:dataset}
\end{table}

\textbf{Image Features:} Previous ZSL approaches use both shallow (SIFT, PHOG, Fisher Vector, color histogram) and deep features \cite{Akata_CVPR_2015,Changpinyo_2016_CVPR,Qiao_2016_CVPR}. As reported repeatedly, deep features outperform shallow features by a significant margin \cite{Changpinyo_2016_CVPR}. For this reason, we consider only deep features from the pretrained GoogLeNet~\cite{GNet_CVPR_2015} and VGG-verydeep-19~\cite{Vgg_arXiv_2014} models for our comparisons. For feature extraction from GoogLeNet and VGG-verydeep-19, we exactly follow  Changpinyo \emph{et al.}~\cite{Changpinyo_2016_CVPR} and Zhang \emph{et al.}~\cite{Zhang_2015_ICCV}, respectively. The dimension of visual features extracted from GoogLeNet is $1024$, and VGG-verydeep-19 is $4096$. While using the recent Xian et al. \cite{Xian_CVPR_2017} seen/unseen split, we use the same 2048-dim features from top-layer pooling units of the 101-layered ResNet \cite{ResNet_CVPR_2016} for a fair comparison.

\textbf{Semantic Space Embeddings:} We analyze both supervised and unsupervised settings of ZSL. For the supervised case, we use 64, 85, 102 and 312 dimensional continuous valued semantic attributes for aPY, AwA, SUN, and CUB datasets, respectively. We dismiss the binary version of these attributes since \cite{Changpinyo_2016_CVPR} showed that continuous attributes are more useful. For the unsupervised case, we test our approach on AwA and CUB datasets. We consider both word vector embeddings i.e., word2vec (w2v) ~\cite{Mikolov_NIPS_2013,Mikolov_arXiv_2013} and GloVe (glo)~\cite{Jeffrey_Glove_2014}. We use $\ell_2$ normalized 400-dimensional word vectors, similar to \cite{Xian_2016_CVPR}. 

\textbf{Evaluation Metrics:} This line of investigation naturally applies to two different tasks; recognition and retrieval~\cite{Xian_CVPR_2017,bucher_ECCV_2016,ReViSE_CoRR_2017}. We measure the recognition performance by the top-1 accuracy, and the retrieval performance by the mean average precision (mAP). The top-1 accuracy is the percentage of the estimated labels (the ones with the highest scores) that match the correct labels. The mean average precision is computed over the precision scores of the test classes. In addition, \cite{Xian_CVPR_2017} proposed to use  Harmonic Mean (HM) of the accuracies of seen and unseen classes ($acc_s$ and $acc_u$ respectively) to evaluate GZSL performance, as follows:
$$HM = \frac{2 \times acc_s \times acc_u}{acc_s + acc_u}.$$
The main motivation of using HM is its ability to estimate the inherent biasness of any method towards seen classes. If a method is too biased to seen classes then $acc_s$ will be very high compared to $acc_u$ and harmonic mean based GZSL performance drops down significantly \cite{Xian_CVPR_2017,Chao_ECCV_2016}.

\textbf{Implementation Details:}\footnote{The code of our method will be released.} We initialize each classifier $\textbf{W}_s$ from a $N(0,\frac{1}{k}$) distribution where $k$ is the dimension of the image feature \cite{Xian_2016_CVPR}.  We use a constant learning rate over 100 iterations in training of each class: $0.001$ for AwA and $0.005$ for aPY, SUN and CUB datasets. For each dataset, we select the value of parameters $\lambda_s$, $\lambda_u$ and $\gamma_s$ using a validation set. We use the same value of $\lambda_s$ and $\lambda_u$ across all seen and unseen classes in the optimization task (Eq.~\ref{eq:seenwights} and \ref{eq:seenwordalpha} respectively). To choose the validation dataset, we divide all seen classes into two groups, and use one group as the unseen set (no test data is used in the validations). Our results on multi-fold cross-validation and single-validation are very similar. 

\subsection{Results for Reduced Set} 
\label{sec:reducedset}

In the reduced set experiment as describe in Sec. \ref{subsec:sparse}, for each unseen class, we select four subsets of the seen classes having one-third of the total number. First three subsets contain the farthest away, mid-range, and nearest seen classes of each unseen class in the semantic embedding space, and the last one is the random selection. For all subsets, we determine the proximity of the unseen classes by Mahalanobis distance with learned metric $\mathbf{M}$. In our experiments, a different unseen class will get a different set of seen classes to describe it. We report the top-1 accuracy on test data of those four subsets in Fig.~\ref{fig:closefar}. We observe that only one-third of seen classes closest to each unseen class perform the best among the four subsets. The farthest away, mid-range and randomly selected subsets fail to describe an unseen class with high accuracy.  This experiment suggest that using only some nearest seen classes located in the semantic embedding space can efficiently approximate an unseen class embedding. The nearest case experiment performances are not the best accuracies reported in the paper because we consider an automatic seen class selection process in our final experiments.

\begin{table}[t]
  \begin{center}
    \begin{tabular}{|l|c|c|c|c|}
    \hline
    Using G & aPY & AwA & CUB & SUN-10 \\ \hline
    Total seen			& 20	&40	&150	&717\\
    Reduced seen-att	&10.17	&20.00	& 74.70	&344.40 \\ 
    Reduced seen-w2v	&-		&21.20	& 70.96		&- \\ 
    Reduced seen-glo	&-		&19.70	& 74.14		&- \\ \hline
    \end{tabular}
     \vspace{-1em}
  \end{center}
  \caption{Average number of the seen classes for reduced set case. Our method automatically selects an optimal number of the nearest seen classes to describe an unseen class.}
  \label{tab:usedseen}
\end{table}

\begin{SCtable*}
  \centering
    \caption{Top-1 accuracy (in \%) of the various versions of CAPD using the attributes. V: VGG-verydeep-19, G: GoogLeNet image features. 
  }
    \begin{tabular}{|l|c|c|c|c|c|c|c|c|}
    \hline
    \multirow{2}{4em}{Method} & \multicolumn{2}{|c|}{aPY} & \multicolumn{2}{|c|}{AwA} & \multicolumn{2}{|c|}{SUN} & \multicolumn{2}{|c|}{CUB}\\ \cline{2-9}
     &V & G & V & G&V & G & V & G\\
    \hline    
    \hline
Ours [all-seen]		&45.84&	50.64&	73.19&	64.74&	84.5&	87.00&	39.86&	42.31\\
Ours [reduced-Lasso]&36.54&	37.22&	74.16&	75.76&	78.50&	84.50&	37.47&	37.37\\
Ours [reduced-auto]	&54.69&	55.07&	78.53&	80.43&	85.00&	79.00   &	43.01&	45.31\\
\hline
\end{tabular}
  \label{tab:ouracc}
\end{SCtable*}

From the discussion in Sec. \ref{subsec:sparse}, we also know that for different unseen classes our method automatically chooses different sets of useful seen classes. The numbers of seen classes in those sets can be different. In Table \ref{tab:usedseen}, we report the average number of seen classes in the sets. One can observe that the average number of the seen classes required is around 50\% across different datasets. This means, in general, only half of the total seen classes are useful to describe one unseen class. Such a reduced set description of the unseen class not only maintains the best performance but also reduces the complexity of the sparse representation of each unseen class.

\begin{figure}[t]
  \begin{center}
  \includegraphics[width=1\linewidth,trim={2.85cm 4.45cm 4cm 5.7cm},clip]{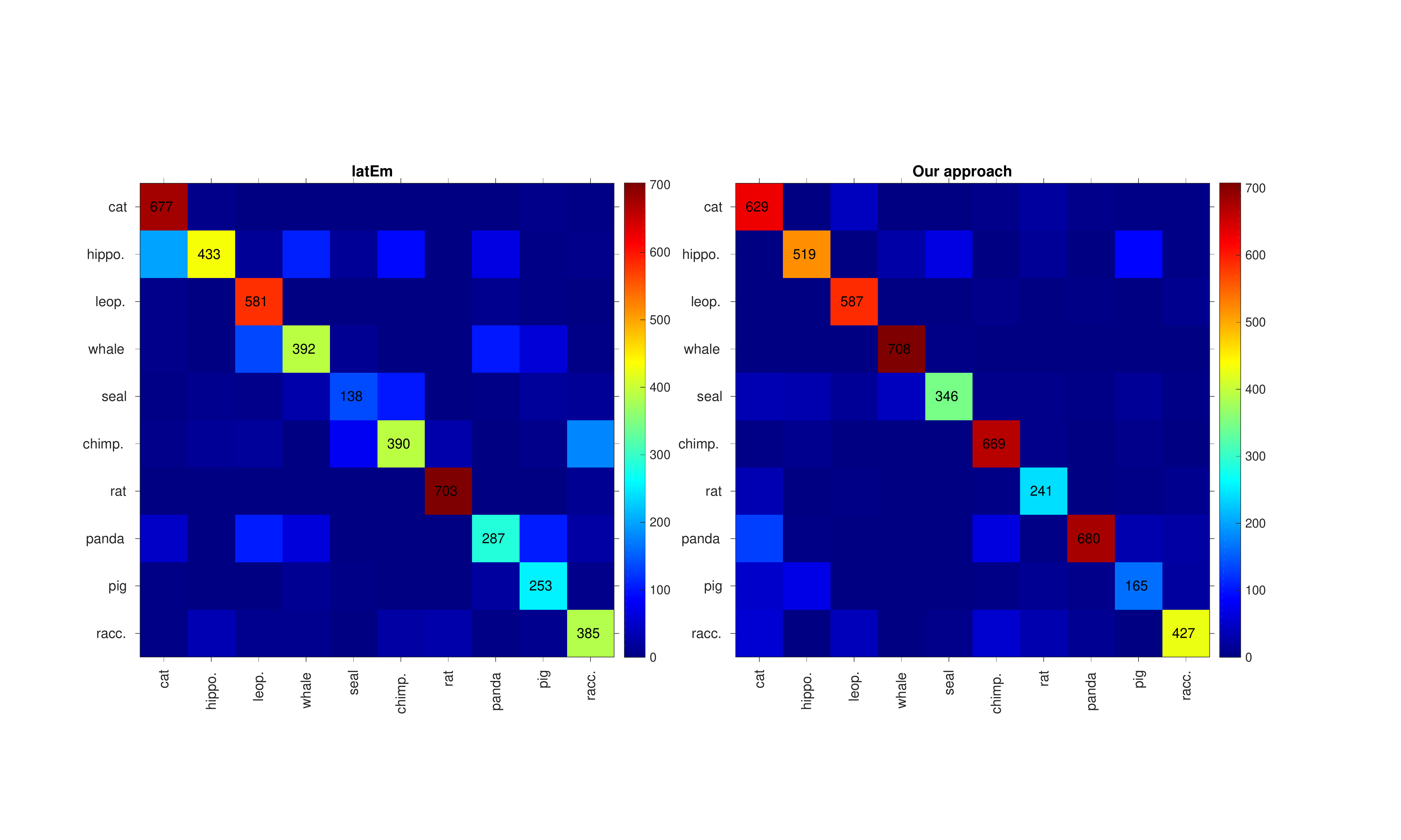}
  \end{center}
   \vspace{-1em}
   \caption{Confusion matrices on AwA dataset using GoogLeNet as image features and the attributes as semantic space vectors. Left: Xian \textit{et al.} \cite{Xian_2016_CVPR}. Right: CAPD. As seen, CAPD provide better overall and class-wise performance.}
\label{fig:confusionMat}
\end{figure}

\begin{table}[t]
  \begin{center}
  
    \begin{tabular}{|l|c|c|c|c|}
    \hline
    Using V &  aPY & AwA & SUN & CUB \\
    \hline\hline
    Lampert'14 \cite{Lampert_PAMI_2014} &  38.16 & 57.23 & 72.00 &31.40\\
    ESZSL'15 \cite{romera_ICML_2015} &  24.22 & 75.32 & 82.10&- \\
    SSE-ReLU'15 \cite{Zhang_2015_ICCV} & 46.23&76.33&82.50&30.41\\
    Zhang'16 \cite{Zhang_2016_CVPR} &  50.35 & \textbf{80.46} & 83.30 &42.11\\
    Bucher'16 \cite{bucher_ECCV_2016} & 53.15 & 77.32 & 84.41 & 43.29\\
    DSRL'17\cite{Ye_DSRL2017_CVPR}& \textbf{56.29} & 77.38 & 82.00 & \textbf{50.26}\\
    MFMR'17\cite{Xu_Matrix_CVPR_2017}  & 48.20 & 79.80 & 84.00 & 47.70\\
    \hline
    Ours & {54.69} & 78.53 &\textbf{85.00} & {43.33} \\
    \hline 
 \end{tabular} 

 \begin{tabular}{|l|c|c|c|c|}   
    \hline
    Using G & aPY & AwA & SUN & CUB \\ \hline \hline
	Lampert'14 \cite{Lampert_PAMI_2014} & 37.10 &59.50&-&-\\
    Akata'15 \cite{Akata_CVPR_2015} & - & 66.70 & - & {50.10}\\
    Changpinyo'16 \cite{Changpinyo_2016_CVPR} & -& 72.90 & - & 45.85\\
    Xian'16 \cite{Xian_2016_CVPR} & -& 72.50 & -& 45.60\\
    SCoRe'17\cite{Morgado_2017_CVPR} & - & 78.30 & - & \textbf{58.40}\\
    MFMR'17\cite{Xu_Matrix_CVPR_2017}  & 46.40 & 76.60 & 81.50 & 46.20\\    
    \hline
    Ours & \textbf{55.07} & \textbf{80.83} & \textbf{87.00} & 45.31\\
    \hline
    \end{tabular}
  \end{center}
  \vspace{-1em}
  \caption{Supervised ZSL top-1 accuracy (in \%) on four standard datasets. V: VGG-verydeep-19 and G: GoogLeNet image features. Results are from the original papers. Only very recent SOTA methods are considered for comparison.}
  \label{tab:otheracc}
\end{table}

\subsection{Benchmark Comparisons \label{comp_zsl}}

We discuss benchmark performances of ZSL recognition and retrieval for both supervised (attributes) and unsupervised semantics (word2vec or GloVe).

\subsubsection{Results for ZSL with Supervised Attributes\protect\footnote{For fairness, inductive test performances from DSRL \cite{Ye_DSRL2017_CVPR}, MFMR \cite{Xu_Matrix_CVPR_2017} and DMaP \cite{Li_2017_CVPR} are reported in the tables.}}
We present the top-1 ZSL accuracy results of different versions of our method in Table \ref{tab:ouracc}. In the all-seen case, we have considered all seen classes to describe an unseen class (Eq.~\ref{eq:uvrelation}). In Lasso, we report the performance using Lasso regularization in place of $\ell_2$ in Eq.~\ref{eq:seenwordalpha}. The results demonstrate that using a reduced number of seen classes to describe an individual unseen class can improve ZSL accuracy. In Table \ref{tab:otheracc}, we compare the overall top-1 accuracy of our method with many recent ZSL approaches. Our approach outperforms other methods in most of the settings. In Fig. \ref{fig:confusionMat}, we show confusion matrices of a recent approach \cite{Xian_2016_CVPR} and ours. Similar to recognition, ZSL can also perform retrieval task. ZSL retrieval is to search images of unseen classes using their class label embeddings. We test the attributes set as a query to retrieve test images. In Table \ref{tab:precision}, we compare our ZSL retrieval performance with four recent approaches on four datasets. Our approach performs consistently better or comparable to state-of-the-art methods.

\begin{table}[t]
  \begin{center}
    \begin{tabular}{|l|c|c|c|c|}
    \hline
    Using V & aPY & AwA & SUN & CUB \\
    \hline\hline
    SSE-INT'15 \cite{Zhang_2015_ICCV} 	  & 15.43	&	46.25	&	58.94	&	4.69\\
    SSE-ReLU'15 \cite{Zhang_2015_ICCV} & 14.09	&	42.60	&	44.55	&	3.70\\
    Bucher'16 \cite{bucher_ECCV_2016} & 36.92	&	68.10	&	52.68	&	25.33\\
    Zhang'16 \cite{Zhang_2016_CVPR}  & 38.30	&	67.66	&	80.10	&	29.15\\ 
    MFMR'17 \cite{Xu_Matrix_CVPR_2017} & \textbf{45.60}	&	70.80	&	77.40	&	30.60\\ 
    \hline    
    Ours & 43.85&\textbf{72.87}&\textbf{80.20}&\textbf{36.60}\\
    \hline
    \end{tabular}   
  \end{center}
  \vspace{-1em}
  \caption{Supervised ZSL retrieval performance (in mAP). V: VGG-verydeep-19 image features.}
  \label{tab:precision}

\end{table}

\begin{table}[t]
  \begin{center}
    \begin{tabular}{|l|c|c|c|c|}
    \hline
    \multirow{2}{8em}{Semantic:word2vec} & \multicolumn{2}{|c|}{AwA} & \multicolumn{2}{|c|}{CUB}\\ \cline{2-5}
     & V & G & V & G\\
    \hline
    Akata'15 \cite{Akata_CVPR_2015} & -&51.20&-&28.40\\    
    Xian'16 \cite{Xian_2016_CVPR} & -&61.10&-&31.80\\
    Akata'16 \cite{Akata_2016_CVPR} & -&-&33.90&-\\
    Changpinyo'16 \cite{Changpinyo_2016_CVPR} &-&57.50&-&-\\ 
	SCoRe'17\cite{Morgado_2017_CVPR} &-&60.88&-&31.51\\
    DMaP-I'17\cite{Li_2017_CVPR} &-&-&-&26.28\\ 
    \hline
    Ours & \textbf{66.26}&\textbf{66.89}&\textbf{34.40}&\textbf{32.42} \\    
    \hline 
 \end{tabular}
 \begin{tabular}{|l|c|c|c|c|}
\hline
   \multirow{2}{8em}{Semantic: GloVe} & \multicolumn{2}{|c|}{AwA} & \multicolumn{2}{|c|}{CUB}\\ \cline{2-5}
     & V & G & V & G\\
    \hline
    Akata'15 \cite{Akata_CVPR_2015} & -&58.80&-&24.20\\
    Xian'16 \cite{Xian_2016_CVPR} & -&62.90&-&\textbf{32.50}\\
    DMaP-I'17\cite{Li_2017_CVPR} &-&-&-&23.69\\ 
    \hline
    Ours  & \textbf{62.01}&\textbf{64.73}& \textbf{32.08} &29.66\\  
	\hline
    \end{tabular}
  \end{center}
  \vspace{-1em}
  \caption{Unsupervised ZSL performance in top-1 accuracy. V: VGG-verydeep-19, G: GoogLeNet image features.
  Only very recent SOTA papers are considered for comparison.}
   \label{tab:unsupervised}
\end{table}

\begin{figure}[t]
  \begin{center}
  \includegraphics[width=.8\linewidth,trim={.7cm .1cm .8cm .1cm},clip]{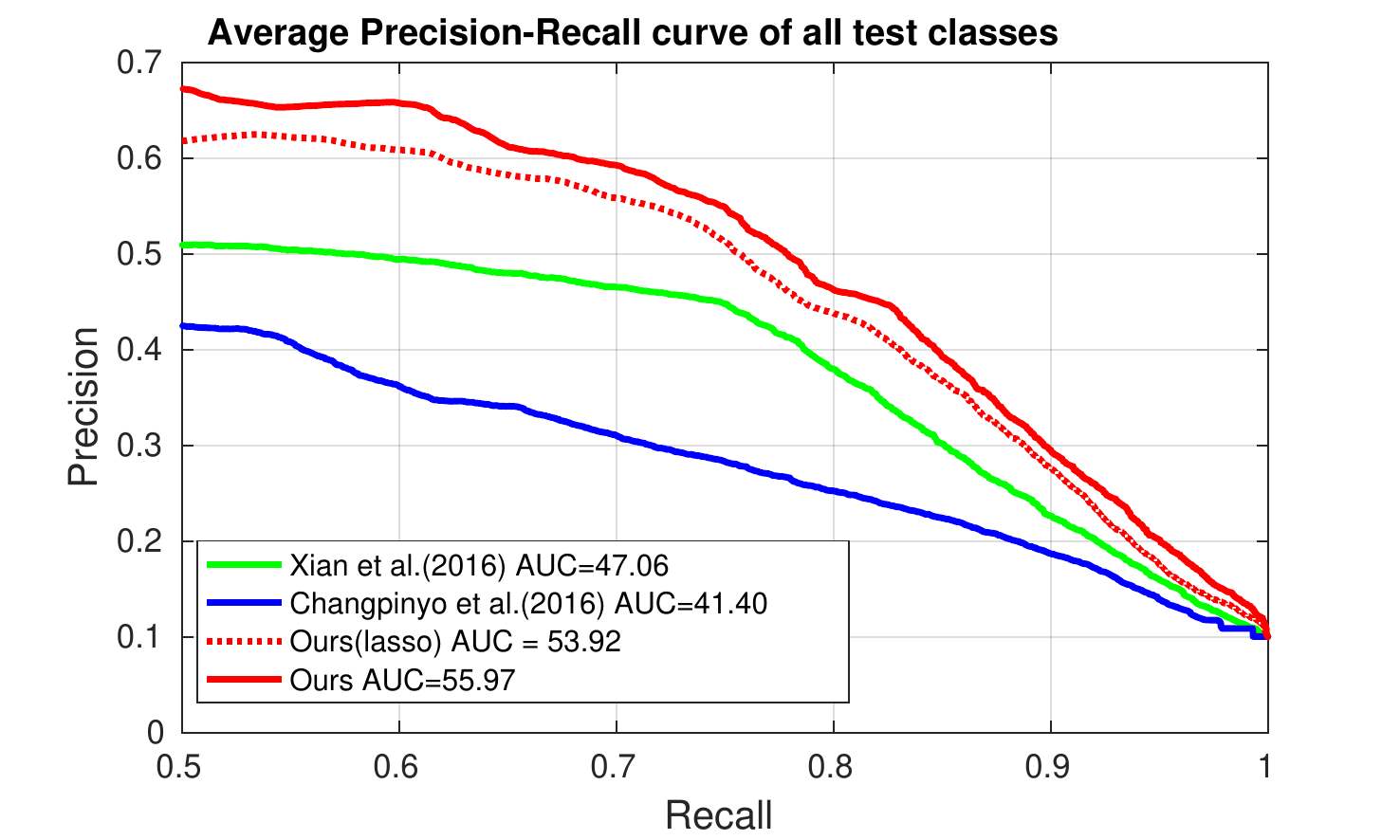}
  \end{center}
  \vspace{-1em}
   \caption{Average precision recall curve of all test classes of AwA dataset. GoogLeNet and word2vec are used as image feature and semantic label embedding respectively.}
\label{fig:prcurve}
\end{figure}

\subsubsection{Results for ZSL with Unsupervised Semantics}
ZSL with pretrained word vectors~\cite{Mikolov_NIPS_2013,Jeffrey_Glove_2014} as sematnic embedding is the focus of attention nowadays since it is difficult to generate manually annotated attribute sets in real-world applications. Therefore, the ZSL research is pushing forward to eliminate dependency on manually assigned attributes \cite{Akata_2016_CVPR,Al-Halah_2016_CVPR,Kodirov_2015_ICCV,Qiao_2016_CVPR,Xian_2016_CVPR}. In line with this view, we adapt our method to unsupervised settings by replacing attribute set with word2vec~\cite{Mikolov_NIPS_2013} and GloVe~\cite{Jeffrey_Glove_2014} vectors. Our results on two standard datasets, AwA and CUB, are reported in Table \ref{tab:unsupervised}.  We compare with very recent approaches keeping same experimental protocol. One can notice that our approach performs consistently in the unsupervised settings as well in a wide variety of feature and semantic embedding combinations. We provide the average precision-recall curves of ours and two very recent approaches using word2vec embeddings in Fig. \ref{fig:prcurve}. As shown, our method is superior to others by a significant margin. 

Our observation is that ZSL attains better performance {with supervised attributes as semantics than unsupervised ones} because the semantic descriptors (word2vec and GloVe) are often noisy and cannot describe a class as good as attributes. To address this performance gap, some works  investigate ZSL with {transductive learning \cite{Xu_Matrix_CVPR_2017,Li_2017_CVPR},} domain adaptation techniques \cite{Elhoseiny_ICCV_2013,Kodirov_2015_ICCV}, and class attribute associations \cite{Akata_2016_CVPR,Al-Halah_2016_CVPR}. In our study, we consider these improvements as future work.

\begin{table*}[t]
  \centering
    \begin{tabular}{|l|ccc|ccc|ccc|ccc|}
    \hline
    \multirow{1}{4em}{Top1} & \multicolumn{3}{|c|}{SUN} & \multicolumn{3}{|c|}{CUB} & \multicolumn{3}{|c|}{AWA} & \multicolumn{3}{|c|}{aPY}\\ \cline{2-13}
    ResNet& HM & $acc_s$ & $acc_u$ & HM & $acc_s$ & $acc_u$ & HM & $acc_s$ & $acc_u$ & HM & $acc_s$ & $acc_u$ \\
    \hline
    DAP\cite{Lampert_PAMI_2014} & 7.2 & 25.1 & 4.2 & 3.3 & 67.9 & 1.7 & 0.0 & \textbf{88.7} & 0.0 & 9.0 & 78.3 & 4.8 \\
    CONSE\cite{norouzi_arXiv_2013} &11.6&39.9&6.8&3.1&\textbf{72.2}&1.6&0.8&88.6&0.4&0.0&\textbf{91.2}&0.0\\
    CMT\cite{Socher_NIPS_2013} &13.3&28.0&8.7&8.7&60.1&4.7&15.3&86.9&8.4&19.0&74.2&10.9\\
    SSE\cite{Zhang_2015_ICCV} &4.0&36.4&2.1&14.4&46.9&8.5&12.9&80.5&7.0&0.4&78.9&0.2\\
    LATEM\cite{Xian_2016_CVPR} &19.5&28.8&14.7&24.0&57.3&15.2&13.3&71.7&7.3&0.2&73.0&0.1\\
    ALE\cite{Akata_PAMI_2016} &26.3&33.1&21.8&34.4&62.8&23.7&27.5&76.1&16.8&8.7&73.7&4.6\\
    DEVISE\cite{DeViSE_NIPS_2013} &20.9&27.4&16.9&32.8&53.0&23.8&22.4&68.7&13.4&9.2&76.9&4.9\\
    SJE\cite{Akata_CVPR_2015} &19.8&30.5&14.7&33.6&59.2&23.5&19.6&74.6&11.3&6.9&55.7&3.7\\
    ESZSL\cite{romera_ICML_2015} &15.8&27.9&11.0&21.0&63.8&12.6&12.1&75.6&6.6&4.6&70.1&2.4\\
    SYNC\cite{Changpinyo_2016_CVPR} &13.4&\textbf{43.3}&7.9&19.8&70.9&11.5&16.2&87.3&8.9&13.3&66.6&7.4\\ \hline
    
    Our GZSL& \textbf{31.3} & 27.8 & \textbf{35.8} & \textbf{43.3} & 41.7 & \textbf{44.9} &\textbf{54.5} & 68.6 & \textbf{45.2} & \textbf{37.0} & 59.5 & \textbf{26.8} \\ \cline{1-13} 
    {Our ZSL} & \multicolumn{3}{|c|}{49.7} & \multicolumn{3}{|c|}{53.8} & \multicolumn{3}{|c|}{52.6} & \multicolumn{3}{|c|}{39.3}\\ \hline
    
 \end{tabular}
 \vspace{-0.4em}
  \caption{GZSL performance comparison with other established methods in the literature. The experiment setting is exactly same as in \cite{Xian_CVPR_2017}. Image features are taken from ResNet and attributes are used as semantic information.}
  \label{tab:GZSL-ResNet}
\end{table*}


\subsubsection{Results for GZSL}
GZSL is a more realistic scenario than conventional ZSL because GZSL setting tests a method with not only the unseen class instances but also seen class instances. In this paper, we extend our method to work on GZSL setting as well. Although GZSL is a more interesting problem than ZSL, usually standard ZSL methods do not report any results on GZSL in the original papers. 
However, recently a few efforts have been published to establish the standard testing protocol for GZSL \cite{Xian_CVPR_2017,Chao_ECCV_2016}. In the current work, we test our GZSL method on both testing protocols of \cite{Xian_CVPR_2017} and \cite{Chao_ECCV_2016}.

Xian et al. \cite{Xian_CVPR_2017} tested 10 ZSL methods with a new seen-unseen split of datasets ensuring unseen classes are not used during pre-training of deep network (e.g., GoogLeNet, ResNet) which was used to extract image features. They used ResNet as image features and attributes as semantic embedding for SUN, CUB, AwA and aPY dataset. With this exact settings, in Table \ref{tab:GZSL-ResNet}, we compare our GZSL results with the reported results of \cite{Xian_CVPR_2017}. In terms of Harmonic based (HM) measure, our results consistently outperform other methods by a large margin. Moreover, our method balances the seen-unseen diversity in a robust manner which helps to achieve the best unseen class accuracy ($acc_u$). In contrast, seen accuracy ($acc_u$) moves down because of the trade-off while balancing the bias towards seen classes. In the last row, we report the ZSL performance of this experiment where only unseen class test instances are classified to only unseen classes (not considering both seen-unseen classes together). This accuracy is actually an oracle case (upper bound) for $acc_u$ of GZSL case of our method. This is because, if an instance is misclassified in the ZSL case, it must be misclassified in the GZSL case too. Another important point to note is that the parameters of our method are tuned for GZSL setting in this experiment. Therefore, ZSL performance in the last row may increase if one tunes parameters for the ZSL setting.

\begin{table}[t]
  \centering
    \begin{tabular}{|l|c|c|c||c|c|c|}
    \hline
    \multirow{2}{4em}{Top1:G} & \multicolumn{3}{|c||}{AwA} & \multicolumn{3}{|c|}{CUB}\\ \cline{2-7}
     & HM & $acc_s$ & $acc_u$ & HM & $acc_s$ & $acc_u$ \\
    \hline
    DAP\cite{Lampert_PAMI_2014}	& 4.7 &77.9	&2.4	& 7.5&55.1	&4.0 \\
    IAP\cite{Lampert_PAMI_2014}	 & 3.3& 76.8 & 1.7 & 2.0& 69.4 & 1.0\\
ConSE\cite{norouzi_arXiv_2013}	&16.9	&75.9	&9.5	&3.5	&69.9	&1.8\\
SynC\cite{Changpinyo_2016_CVPR} &	0.8&\textbf{81.0}&	0.4&	22.3&\textbf{72.0}&	13.2 \\
MFMR\cite{Xu_Matrix_CVPR_2017}	&29.60&75.6	&18.4	&-&-&-\\
\hline
Our GZSL &\textbf{50.8}&43.2&\textbf{61.7}&\textbf{29.5}&23.4&\textbf{39.9} \\ \hline
Our ZSL& \multicolumn{3}{|c||}{76.2} & \multicolumn{3}{|c|}{44.0}\\ \hline
 \end{tabular}
 \vspace{-0.5em}
 \caption{GZSL performance comparison with the experiment settings of \cite{Chao_ECCV_2016}. Image features are taken from GoogLeNet and attributes are used as semantic information.}
 \label{tab:GZSL-GNet}
\end{table}

\begin{table}[t]
  \centering
    \begin{tabular}{|l|c|c|c|c|}
    \hline
     \multirow{2}{4em}{Top1:Mean} & \multicolumn{2}{|c|}{AwA} & \multicolumn{2}{|c|}{CUB}\\ \cline{2-5}
      & att & w2v & att & glo \\
    \hline
     DMaP\cite{Li_2017_CVPR}	& 17.23 & 6.44	& 13.55	& 2.07\\
     Our GZSL &\textbf{52.45}&\textbf{43.70}&\textbf{31.65}&\textbf{18.75} \\ \hline
 \end{tabular}
 \vspace{-0.5em}
 \caption{Comparison with a recent GZSL work DMaP\cite{Li_2017_CVPR}}
 \label{tab:GZSL-DMaP}
\end{table}

Chao et al. \cite{Chao_ECCV_2016} experimented GZSL with standard seen-unseen split used in ZSL literature. Keeping this split, they kept random 80\% seen class images for training and held out the rest of 20\% images for testing stage during GZSL. We perform the same harmonic mean based evaluation like previous setting. In Table \ref{tab:GZSL-GNet}, we compare our results with the reported results in \cite{Chao_ECCV_2016}. {Using the same settings, we also compare with two recent methods, MFMR \cite{Xu_Matrix_CVPR_2017} (Table \ref{tab:GZSL-GNet}) and DMAP \cite{Li_2017_CVPR} (Table \ref{tab:GZSL-DMaP}). For the sake of comparison with DMAP \cite{Li_2017_CVPR}, we compare mean Top1 accuracy (not standard though) instead of harmonic mean because $acc_u$ and $acc_s$ are not reported separately in the \cite{Li_2017_CVPR}. Again, our method performs consistently well across datasets.} More results on GZSL for AwA, CUB, SUN and aPY datasets are reported in Tables \ref{tab:AwA_allresult}, \ref{tab:CUB_allresult}, \ref{tab:SUN_allresult} and \ref{tab:aPY_allresult}.

\subsubsection{Results for FSL}
As stated earlier, our method can easily take the advantage when new unseen class instances become available as labeled data for training. To test this scenario, in FSL settings, we assume three instances of each unseen class (randomly chosen) are available as labeled during training. In Table \ref{tab:FSL-comparison}, we report our results for FSL on AwA and CUB dataset while using attribute, word2vec and GloVe as semantic information. The compared methods, DeViSE \cite{Frome_NIPS_2013} and CMT\cite{Socher_NIPS_2013}, did not report FSL performance in the original paper. But, \cite{ReViSE_CoRR_2017} reimplemented the original work to adapt FSL. The exact three instances of each unseen class used in \cite{ReViSE_CoRR_2017} are not publically available. However, to make our results comparable with others, we report the average performance of 10 random trails. Our method performs consistently better than comparing methods except one case: mAP of CUB-att (58.0 vs 58.5). Another observation from these results is that the performance gap between unsupervised semantics (like word2vec and GloVe) and supervised attribute semantics is significantly reduced compared to ZSL settings where unsupervised semantics always ill-performed than supervised attributes across all methods. The reason is that the FSL setting alleviates the inherent noise of unsupervised semantics to perform better (and as good as) supervised semantic. We also experiment on the OSL task, where all conditions are same as FSL setting except a single randomly picked labeled instance is available for each unseen class during training. More results of OSL and FSL for AwA, CUB, SUN and aPY datasets are reported in Table \ref{tab:AwA_allresult}, \ref{tab:CUB_allresult}, \ref{tab:SUN_allresult} and \ref{tab:aPY_allresult}.

\begin{figure}[t]
  \begin{center}
  \includegraphics[width=1\linewidth,trim={2.6cm .1cm 3cm .4cm},clip]{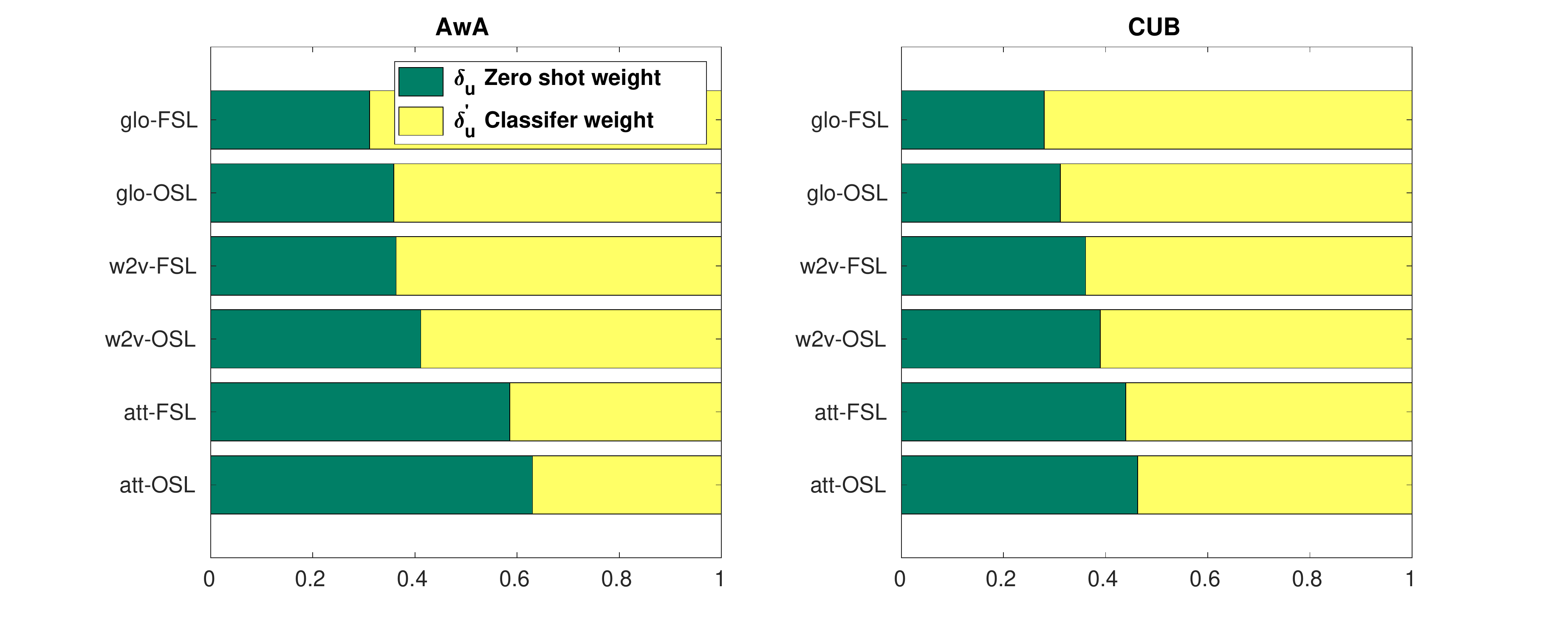}
  \end{center}
  \vspace{-1.3em}
   \caption{Contribution of $\delta_u$ and $\delta_u^{'}$ to update unseen class CAPD}
\label{fig:fewshotbar}
\end{figure}

For any given image, our FSL method described in Sect.~\ref{subsec:FSL} utilizes the contribution of unseen CAPDs coming from two sources: one by combining the CAPDs of seen classes from zero-shot setting and another by using unseen classifier from few-shot setting. In Eq. \ref{eq:fewshotcombination}, two constants ($\delta_u$ and $\delta_u^{'}$) combine the respective CAPDs to compute the updated CAPD of the unseen class. In this experiment, we visualize the contribution of $\delta_u$ and $\delta_u^{'}$ for AwA and CUB dataset in Fig. \ref{fig:fewshotbar}. Few observations from this figure are below:
\begin{itemize}
\item In most cases, few-shot contribution from classifier ($\delta_u^{'}$) contributes higher than zero-shot contribution ($\delta_u$). The reason is that few instances of unseen class can make better generalization than no instance during training.

\item Zero-shot contribution ($\delta_u$) contributes higher on supervised attribute case than word2vec or GloVe across two datasets. The reason is that supervised attributes contain less noise which gives high confidence to zero-shot based CAPD.

\item While comparing OSL and FSL, few-shot contribution  from classifier ($\delta_u^{'}$) contributes higher in FSL than OSL case. The reason is that in FSL settings, any unseen classifier becomes more confident than OSL settings as FSL observes more than one instances during training.

\item While comparing word2vec and GloVe for both OSL and FSL settings, zero-shot contribution ($\delta_u$) contributes higher for word2vec than GloVe semantics. It suggests that word2vec is a better semantic embedding than GloVe for FSL task.

\item While comparing AwA and CUB, zero-shot contribution ($\delta_u$) contributes lower than few-shot contribution from classifier ($\delta_u^{'}$) for CUB across all semantics used. The reason is that CUB is a more difficult dataset than AwA in zero-shot setting. One can find that the overall performance on CUB is lower than AwA in all cases (i.e., ZSL, F/OSL and GZSL).
\end{itemize}

\begin{table}[t]
  \centering
    \begin{tabular}{|l|c|c|c||c|c|c|}
    \hline
    \multirow{2}{5em}{Top1: Using G} & \multicolumn{3}{|c||}{AwA} & \multicolumn{3}{|c|}{CUB}\\ \cline{2-7}
     & att & w2v & glo & att & w2v & glo \\
    \hline
    DeViSE\cite{Frome_NIPS_2013} &80.9&75.3&79.4&54.0&45.7&46.0 \\
    CMT\cite{Socher_NIPS_2013} &85.1&83.4&84.3&56.7&53.4&52.0 \\
    Our &\textbf{87.4}&\textbf{84.9}&\textbf{85.8}&\textbf{56.9}&\textbf{55.4}&\textbf{55.8} \\
    \hline
 \end{tabular}
 
   \begin{tabular}{|l|c|c|c||c|c|c|}
    \hline
    \multirow{2}{5em}{mAP: Using G} & \multicolumn{3}{|c||}{AwA} & \multicolumn{3}{|c|}{CUB}\\ \cline{2-7}
     & att & w2v & glo & att & w2v & glo \\
    \hline
    DeViSE\cite{Frome_NIPS_2013} &85.0&79.3&84.9&46.4&42.6&42.9 \\
    CMT\cite{Socher_NIPS_2013} &88.4&88.2&89.2&\textbf{58.5}&54.0&52.7 \\
    Our &\textbf{92.0}&\textbf{89.5}&\textbf{89.6}&58.0&\textbf{56.3}&\textbf{56.2} \\
    \hline 
 \end{tabular}
 \vspace{-0.6em}
 \caption{FSL performance comparison with the experiment settings of \cite{ReViSE_CoRR_2017}. Image features are taken from GoogLeNet.}
 \label{tab:FSL-comparison}
\end{table}

\subsubsection{All results at a glance.}
With experiment setting of \cite{Chao_ECCV_2016}, we juxtapose all results of OSL, FSL, ZSL and GZSL for AwA, CUB, SUN and aPY datasets in Table \ref{tab:AwA_allresult}, \ref{tab:CUB_allresult}, \ref{tab:SUN_allresult} and \ref{tab:aPY_allresult} respectively. Some overall observations from these results are below:

\begin{itemize}

\item Performance improves from OSL to FSL settings. This is expected because in FSL setting, more than one (three to be exact) instances of unseen class are used as labeled during training.

\item The performance gap between supervised attributes and unsupervised word2vec or GloVe is greatly reduced in OSL and FSL. It suggests that getting few instances as labeled during training helps to greatly compensate the noise of unsupervised semantics.

\item O/FSL setting should always outperform ZSL because more information of unseen is revealed in O/FSL settings. However, we got one exception in SUN dataset where OSL perform worse than ZSL. The reason is that the SUN dataset has 717 classes and only one labeled instance of unseen class could not provide discriminative information which eventually confuses our auto unseen CAPD weighting process.

\item ZSL results are different from Table \ref{tab:otheracc}, \ref{tab:precision} and \ref{tab:unsupervised} because here our method is tuned for GZSL case not on ZSL. In addition, random selection of 80\% training instance of seen classes across 10 different trails affects the result.

\item Performance of $acc_u$ of GZSL is always lower than ZSL because ZSL accuracy is the oracle case of $acc_u$.

\end{itemize}

\begin{table}[t]
  \centering
    \begin{tabular}{|l|c|c||c|c|c|c|c|}
    \hline
     Using G &\multirow{2}{2em}{OSL}&\multirow{2}{2em}{FSL}&\multirow{2}{2em}{ZSL}&\multicolumn{3}{|c|}{GZSL} \\ \cline{5-7}
      Semantic&&&&HM & $acc_s$ & $acc_u$ \\ \hline
     Top1:att&71.2&83.6&40.7&35.7&40.5&32.0 \\
     mAP: att&77.7&88.3&45.1&27.7&24.1&32.7 \\
    \hline
 \end{tabular}
 \vspace{-0.7em}
 \caption{All results on aPY dataset at a glance.}
 \label{tab:aPY_allresult}
\end{table}

\begin{table}[t]
  \centering
    \begin{tabular}{|l|c|c||c|c|c|c|c|}
    \hline
     Using G &\multirow{2}{2em}{OSL}&\multirow{2}{2em}{FSL}&\multirow{2}{2em}{ZSL}&\multicolumn{3}{|c|}{GZSL} \\ \cline{5-7}
      Semantic&&&&HM & $acc_s$ & $acc_u$ \\ \hline
     Top1:att&82.8&87.4&76.2&50.8&43.2&61.7 \\
     mAP: att&86.9&92.0&71.7&50.0&41.2&63.6 \\ \hline
     Top1:w2v&76.9&84.7&56.4&43.6&42.8&44.6 \\
     mAP: w2v&82.0&89.5&50.8&38.5&35.3&42.5 \\ \hline
     Top1:glo&78.2&85.8&60.7&44.7&46.4&43.2 \\
     mAP: glo&83.7&89.6&54.3&42.2&37.8&47.9 \\ \hline
 \end{tabular}
 \vspace{-0.7em}
 \caption{All results on AwA dataset at a glance.}
 \label{tab:AwA_allresult}
\end{table}

\begin{table}[t]
  \centering
    \begin{tabular}{|l|c|c||c|c|c|c|c|}
    \hline
     Using G &\multirow{2}{2em}{OSL}&\multirow{2}{2em}{FSL}&\multirow{2}{2em}{ZSL}&\multicolumn{3}{|c|}{GZSL} \\ \cline{5-7}
      Semantic&&&&HM & $acc_s$ & $acc_u$ \\ \hline
     Top1:att&46.3&56.9&44.0&29.5&23.4&39.9 \\
     mAP: att&46.9&58.0&40.5&31.8&29.2&34.9 \\ \hline
     Top1:w2v&41.7&55.4&33.2&14.9&9.8&31.1 \\
     mAP: w2v&41.9&56.3&29.5&23.2&21.9&24.6 \\ \hline
     Top1:glo&41.2&55.8&31.1&11.7&7.2&30.3 \\
     mAP: glo&40.3&56.2&28.3&23.1&22.8&23.4 \\ \hline
 \end{tabular}
  \vspace{-0.7em}
 \caption{All results on CUB dataset at a glance.}
 \label{tab:CUB_allresult}
\end{table}

\begin{table}[t]
  \centering
    \begin{tabular}{|l|c|c||c|c|c|c|c|}
    \hline
     Using G &\multirow{2}{2em}{OSL}&\multirow{2}{2em}{FSL}&\multirow{2}{2em}{ZSL}&\multicolumn{3}{|c|}{GZSL} \\ \cline{5-7}
     Semantic&&&&HM & $acc_s$ & $acc_u$ \\ \hline
     \multicolumn{7}{|c|}{SUN (645/72: Seen/Unseen Split)}\\ \hline
     Top1:att&53.7&66.3&59.8&28.3&22.2&39.2 \\
     mAP: att&55.2&68.9&60.5&34.1&27.1&45.9 \\ \hline
     \multicolumn{7}{|c|}{SUN-10 (707/10: Seen/Unseen Split)}\\ \hline
     Top1:att&80.8&87.5&77.9&33.6&25.7&48.6 \\
     mAP: att&84.3&90.1&76.8&40.0&32.3&52.7 \\ \hline
 \end{tabular}
  \vspace{-0.7em}
 \caption{All results on SUN dataset at a glance.}
 \label{tab:SUN_allresult}
\end{table}

\subsection{Discussion}

Based on our experiments, we draw the following contributions of our work:

\textbf{Benefits of CAPD:} A CAPD points out the most likely class. If a semantic space embedding vector of a class and the CAPD of the image lies close to each other, there is a strong confidence for that class. One important contribution of this paper is the derivation of the CAPD for each unseen class. Conventional ZSL approaches in this vein of thought essentially calculate one principal direction \cite{Akata_CVPR_2015,romera_ICML_2015,Xian_2016_CVPR,Qiao_2016_CVPR,Mubarak_2016_CVPR}. Generalizing all seen-unseen classes with only one principal direction cannot capture the differences among classes effectively. In our work, each CAPD is obtained with the help of bilinear mapping (matrix multiplication). One can extend this by incorporating latent variables, in line with the work Xian et al. \cite{Xian_2016_CVPR} where a collection of bilinear maps along with a selection criterion is used. 


\textbf{Benefits of Nearest Seen Classes:} Intuitively, when we describe a novel object, rather than giving a dissimilar object as an example, we use a similar known object. This hints that we can reconstruct the CAPD of an unseen class with the CAPDs of the similar seen classes. This idea helps to improve the prediction performance. 

\textbf{How Many Seen Classes are Required?} Results presented in Fig.~\ref{fig:closefar} support the idea that all seen classes are not always necessary. We propose a simple yet effective solution for selecting adaptively the number of similar seen classes for each unseen class (see the discussion in Sec.~\ref{subsec:sparse}). This scheme allows different set of useful seen classes required to describe an unseen class.

\textbf{Extension to GZSL Setting:} ZSL methods are biased to assign high prediction scores towards seen classes while performing GZSL task. Due to this reason, conventional ZSL methods fail to achieve good performance in GZSL. Our proposed method solves this problem by adapting seen-unseen class diversity in a novel manner. Unlike \cite{Socher_NIPS_2013,Chao_ECCV_2016}, our adaptation technique does not take any extra supervision from training/validation image data. We show that class semantic information can be used to adapt seen-unseen diversity.

\textbf{Extension to Few/One Shot Settings:} In some applications, a few images of a new class may become available for training. To adapt with such situations, our method can train a model for the new class without disturbing the previous training. The CAPD from the new model is combined with its previous CAPD (of unseen settings) to obtain an updated CAPD with few-shot refinement. We propose an automatic way of combining CAPDs from two sources by measuring the quality of prediction responses of training images. Our updated CAPD provides better fitness score for unseen class prediction.

\section{Conclusion}

We propose a novel unified solution to ZSL, GZSL and F/OSL problems utilizing the concept of class adaptive principal direction (CAPD) that enables efficient and discriminative embeddings of unseen class images in semantic space for recognition and retrieval. We introduce an automatic solution to select a reduced set of relevant seen classes. As demonstrated in our extensive experimental analysis, our method works consistently well in both unsupervised and supervised ZSL settings and achieves the superior performance in particular for the unsupervised case. It provides several benefits including reliable generalization and noise suppression in the semantic space. In addition to ZSL, our method also performs very well in GZSL settings. We propose an easy solution to match the seen-unseen diversity of classes at the algorithmic level. Unlike conventional methods, our GZSL strategy can balance seen-unseen performance to achieve overall better recognition rates. We have extended our CAPD based ZSL approach to adapt with FSL settings. Our approach easily takes the advantage of few examples available in FSL task to fine tune unseen CAPDs to improve classification performance. As a future work, we will extend our approach to transductive settings and domain adaptation.


\ifCLASSOPTIONcaptionsoff
  \newpage
\fi


{\small
\bibliographystyle{ieee}
\bibliography{ref}

\begin{thebibliography}{10}\itemsep=-1pt

\bibitem{Akata_2016_CVPR}
Z.~Akata, M.~Malinowski, M.~Fritz, and B.~Schiele.
\newblock Multi-cue zero-shot learning with strong supervision.
\newblock In {\em The IEEE Conference on Computer Vision and Pattern
  Recognition (CVPR)}, June 2016.

\bibitem{Akata_CVPR_2013}
Z.~Akata, F.~Perronnin, Z.~Harchaoui, and C.~Schmid.
\newblock Label-embedding for attribute-based classification.
\newblock In {\em Proceedings of the IEEE Computer Society Conference on
  Computer Vision and Pattern Recognition}, pages 819--826, 2013.

\bibitem{Akata_PAMI_2016}
Z.~Akata, F.~Perronnin, Z.~Harchaoui, and C.~Schmid.
\newblock {Label-Embedding for Image Classification}.
\newblock {\em {IEEE Transactions on Pattern Analysis and Machine
  Intelligence}}, 38(7):1425--1438, July 2016.

\bibitem{Akata_CVPR_2015}
Z.~Akata, S.~Reed, D.~Walter, H.~Lee, and B.~Schiele.
\newblock Evaluation of output embeddings for fine-grained image
  classification.
\newblock In {\em Proceedings of the IEEE Computer Society Conference on
  Computer Vision and Pattern Recognition}, volume 07-12-June-2015, pages
  2927--2936, 2015.

\bibitem{Al-Halah_2016_CVPR}
Z.~Al-Halah, M.~Tapaswi, and R.~Stiefelhagen.
\newblock Recovering the missing link: Predicting class-attribute associations
  for unsupervised zero-shot learning.
\newblock In {\em The IEEE Conference on Computer Vision and Pattern
  Recognition (CVPR)}, June 2016.

\bibitem{Bendale_CVPR_2016}
A.~Bendale and T.~E. Boult.
\newblock Towards open set deep networks.
\newblock In {\em Proceedings of the IEEE Conference on Computer Vision and
  Pattern Recognition}, pages 1563--1572, 2016.

\bibitem{Changpinyo_2016_CVPR}
S.~Changpinyo, W.-L. Chao, B.~Gong, and F.~Sha.
\newblock Synthesized classifiers for zero-shot learning.
\newblock In {\em Proceedings of the IEEE Computer Society Conference on
  Computer Vision and Pattern Recognition}, volume 2016-January, pages
  5327--5336, 2016.

\bibitem{Changpinyo_2017_ICCV}
S.~Changpinyo, W.-L. Chao, and F.~Sha.
\newblock Predicting visual exemplars of unseen classes for zero-shot learning.
\newblock In {\em The IEEE International Conference on Computer Vision (ICCV)},
  Oct 2017.

\bibitem{Chao_ECCV_2016}
W.-L. Chao, B.~Changpinyo, Soravitand~Gong, and F.~Sha.
\newblock {\em An Empirical Study and Analysis of Generalized Zero-Shot
  Learning for Object Recognition in the Wild}, pages 52--68.
\newblock Springer International Publishing, Cham, 2016.

\bibitem{Elhoseiny_ICCV_2013}
M.~Elhoseiny, B.~Saleh, and A.~Elgammal.
\newblock Write a classifier: Zero-shot learning using purely textual
  descriptions.
\newblock In {\em Proceedings of the IEEE International Conference on Computer
  Vision}, pages 2584--2591, 2013.

\bibitem{aPY_2009}
A.~Farhadi, I.~Endres, D.~Hoiem, and D.~Forsyth.
\newblock Describing objects by their attributes.
\newblock In {\em Computer Vision and Pattern Recognition, 2009. CVPR 2009.
  IEEE Conference on}, pages 1778--1785. IEEE, 2009.

\bibitem{Fei_PAMI_2006}
L.~Fei-Fei, R.~Fergus, and P.~Perona.
\newblock One-shot learning of object categories.
\newblock {\em IEEE Transactions on Pattern Analysis and Machine Intelligence},
  28(4):594--611, April 2006.

\bibitem{Frome_NIPS_2013}
A.~Frome, G.~S. Corrado, J.~Shlens, S.~Bengio, J.~Dean, M.~Ranzato, and
  T.~Mikolov.
\newblock Devise: A deep visual-semantic embedding model.
\newblock In {\em NIPS}, 2013.

\bibitem{DeViSE_NIPS_2013}
A.~Frome, G.~S. Corrado, J.~Shlens, S.~Bengio, J.~Dean, M.~A. Ranzato, and
  T.~Mikolov.
\newblock Devise: A deep visual-semantic embedding model.
\newblock In C.~J.~C. Burges, L.~Bottou, M.~Welling, Z.~Ghahramani, and K.~Q.
  Weinberger, editors, {\em Advances in Neural Information Processing Systems
  26}, pages 2121--2129. Curran Associates, Inc., 2013.

\bibitem{Gavves_ICCV_2015}
E.~Gavves, T.~Mensink, T.~Tommasi, C.~G.~M. Snoek, and T.~Tuytelaars.
\newblock Active transfer learning with zero-shot priors: Reusing past datasets
  for future tasks.
\newblock In {\em The IEEE International Conference on Computer Vision (ICCV)},
  December 2015.

\bibitem{ResNet_CVPR_2016}
K.~He, X.~Zhang, S.~Ren, and J.~Sun.
\newblock Deep residual learning for image recognition.
\newblock volume 2016-January, pages 770--778, 2016.
\newblock cited By 107.

\bibitem{Jain_ECCV_2014}
L.~P. Jain, W.~J. Scheirer, and T.~E. Boult.
\newblock Multi-class open set recognition using probability of inclusion.
\newblock In {\em European Conference on Computer Vision}, pages 393--409.
  Springer, 2014.

\bibitem{Jayaraman_NIPS_2014}
D.~Jayaraman and K.~Grauman.
\newblock Zero-shot recognition with unreliable attributes.
\newblock In Z.~Ghahramani, M.~Welling, C.~Cortes, N.~D. Lawrence, and K.~Q.
  Weinberger, editors, {\em Advances in Neural Information Processing Systems
  27}, pages 3464--3472. Curran Associates, Inc., 2014.

\bibitem{Kodirov_2015_ICCV}
E.~Kodirov, T.~Xiang, Z.~Fu, and S.~Gong.
\newblock Unsupervised domain adaptation for zero-shot learning.
\newblock In {\em The IEEE International Conference on Computer Vision (ICCV)},
  December 2015.

\bibitem{Krause_CVPR_2015}
J.~Krause, H.~Jin, J.~Yang, and L.~Fei-Fei.
\newblock Fine-grained recognition without part annotations.
\newblock In {\em Proceedings of the IEEE Conference on Computer Vision and
  Pattern Recognition}, pages 5546--5555, 2015.

\bibitem{AwA_2009}
C.~Lampert, H.~Nickisch, and S.~Harmeling.
\newblock Learning to detect unseen object classes by between-class attribute
  transfer.
\newblock In {\em 2009 IEEE Computer Society Conference on Computer Vision and
  Pattern Recognition Workshops, CVPR Workshops 2009}, pages 951--958, 2009.

\bibitem{Lampert_PAMI_2014}
C.~H. Lampert, H.~Nickisch, and S.~Harmeling.
\newblock Attribute-based classification for zero-shot visual object
  categorization.
\newblock {\em IEEE Transactions on Pattern Analysis and Machine Intelligence},
  36(3):453--465, March 2014.

\bibitem{Li_2017_CVPR}
Y.~Li, D.~Wang, H.~Hu, Y.~Lin, and Y.~Zhuang.
\newblock Zero-shot recognition using dual visual-semantic mapping paths.
\newblock In {\em The IEEE Conference on Computer Vision and Pattern
  Recognition (CVPR)}, July 2017.

\bibitem{bucher_ECCV_2016}
S.~H. Maxime~Bucher and F.~Jurie.
\newblock Improving semantic embedding consistency by metric learning for
  zero-shot classification.
\newblock In {\em Proceedings of The 14th European Conference on Computer
  Vision}, 2016.

\bibitem{Mensink_CVPR_2014}
T.~Mensink, E.~Gavves, and C.~G. Snoek.
\newblock Costa: Co-occurrence statistics for zero-shot classification.
\newblock In {\em Proceedings of the IEEE Conference on Computer Vision and
  Pattern Recognition}, pages 2441--2448, 2014.

\bibitem{Mikolov_arXiv_2013}
T.~Mikolov, K.~Chen, G.~Corrado, and J.~Dean.
\newblock Efficient estimation of word representations in vector space.
\newblock {\em arXiv preprint arXiv:1301.3781}, January 2013.

\bibitem{Mikolov_NIPS_2013}
T.~Mikolov, I.~Sutskever, K.~Chen, G.~S. Corrado, and J.~Dean.
\newblock Distributed representations of words and phrases and their
  compositionality.
\newblock In C.~J.~C. Burges, L.~Bottou, M.~Welling, Z.~Ghahramani, and K.~Q.
  Weinberger, editors, {\em Advances in Neural Information Processing Systems
  26}, pages 3111--3119. Curran Associates, Inc., 2013.

\bibitem{Morgado_2017_CVPR}
P.~Morgado and N.~Vasconcelos.
\newblock Semantically consistent regularization for zero-shot recognition.
\newblock In {\em The IEEE Conference on Computer Vision and Pattern
  Recognition (CVPR)}, July 2017.

\bibitem{norouzi_arXiv_2013}
M.~Norouzi, T.~Mikolov, S.~Bengio, Y.~Singer, J.~Shlens, A.~Frome, G.~S.
  Corrado, and J.~Dean.
\newblock Zero-shot learning by convex combination of semantic embeddings.
\newblock {\em arXiv preprint arXiv:1312.5650}, 2013.

\bibitem{Hinton_NIPS_2009}
M.~Palatucci, D.~Pomerleau, G.~E. Hinton, and T.~M. Mitchell.
\newblock Zero-shot learning with semantic output codes.
\newblock In Y.~Bengio, D.~Schuurmans, J.~D. Lafferty, C.~K.~I. Williams, and
  A.~Culotta, editors, {\em Advances in Neural Information Processing Systems
  22}, pages 1410--1418. Curran Associates, Inc., 2009.

\bibitem{SUN_2014}
G.~Patterson, C.~Xu, H.~Su, and J.~Hays.
\newblock The sun attribute database: Beyond categories for deeper scene
  understanding.
\newblock {\em International Journal of Computer Vision}, 108(1-2):59--81,
  2014.

\bibitem{Jeffrey_Glove_2014}
J.~Pennington, R.~Socher, and C.~D. Manning.
\newblock Glove: Global vectors for word representation.
\newblock In {\em Empirical Methods in Natural Language Processing (EMNLP)},
  pages 1532--1543, 2014.

\bibitem{Qiao_2016_CVPR}
R.~Qiao, L.~Liu, C.~Shen, and A.~van~den Hengel.
\newblock Less is more: Zero-shot learning from online textual documents with
  noise suppression.
\newblock In {\em The IEEE Conference on Computer Vision and Pattern
  Recognition (CVPR)}, June 2016.

\bibitem{Rohrbach_CVPR_2011}
M.~Rohrbach, M.~Stark, and B.~Schiele.
\newblock Evaluating knowledge transfer and zero-shot learning in a large-scale
  setting.
\newblock In {\em Computer Vision and Pattern Recognition (CVPR), 2011 IEEE
  Conference on}, pages 1641--1648. IEEE, 2011.

\bibitem{romera_ICML_2015}
B.~Romera-Paredes and P.~Torr.
\newblock An embarrassingly simple approach to zero-shot learning.
\newblock In {\em Proceedings of The 32nd International Conference on Machine
  Learning}, pages 2152--2161, 2015.

\bibitem{Salakhutdinov_PAMI_2013}
R.~Salakhutdinov, J.~B. Tenenbaum, and A.~Torralba.
\newblock Learning with hierarchical-deep models.
\newblock {\em IEEE Transactions on Pattern Analysis and Machine Intelligence},
  35(8):1958--1971, Aug 2013.

\bibitem{Vgg_arXiv_2014}
K.~Simonyan and A.~Zisserman.
\newblock Very deep convolutional networks for large-scale image recognition.
\newblock {\em arXiv preprint arXiv:1409.1556}, 2014.

\bibitem{Socher_NIPS_2013}
R.~Socher, M.~Ganjoo, C.~D. Manning, and A.~Ng.
\newblock Zero-shot learning through cross-modal transfer.
\newblock In C.~J.~C. Burges, L.~Bottou, M.~Welling, Z.~Ghahramani, and K.~Q.
  Weinberger, editors, {\em Advances in Neural Information Processing Systems
  26}, pages 935--943. Curran Associates, Inc., 2013.

\bibitem{GNet_CVPR_2015}
C.~Szegedy, W.~Liu, Y.~Jia, P.~Sermanet, S.~Reed, D.~Anguelov, D.~Erhan,
  V.~Vanhoucke, and A.~Rabinovich.
\newblock Going deeper with convolutions.
\newblock {\em Proceedings of the IEEE Computer Society Conference on Computer
  Vision and Pattern Recognition}, 07-12-June-2015:1--9, 2015.

\bibitem{ReViSE_CoRR_2017}
Y.~H. Tsai, L.~Huang, and R.~Salakhutdinov.
\newblock Learning robust visual-semantic embeddings.
\newblock {\em CoRR}, abs/1703.05908, 2017.

\bibitem{tSNE_van2014}
L.~Van Der~Maaten.
\newblock Accelerating t-sne using tree-based algorithms.
\newblock {\em Journal of machine learning research}, 15(1):3221--3245, 2014.

\bibitem{CUB_2011}
C.~Wah, S.~Branson, P.~Welinder, P.~Perona, and S.~Belongie.
\newblock {The Caltech-UCSD Birds-200-2011 Dataset}.
\newblock Technical Report CNS-TR-2011-001, California Institute of Technology,
  2011.

\bibitem{Wang_CVPR_2013}
X.~Wang and Q.~Ji.
\newblock A unified probabilistic approach modeling relationships between
  attributes and objects.
\newblock {\em Proceedings of the IEEE International Conference on Computer
  Vision}, pages 2120--2127, 2013.

\bibitem{Xian_2016_CVPR}
Y.~Xian, Z.~Akata, G.~Sharma, Q.~Nguyen, M.~Hein, and B.~Schiele.
\newblock Latent embeddings for zero-shot classification.
\newblock In {\em The IEEE Conference on Computer Vision and Pattern
  Recognition (CVPR)}, June 2016.

\bibitem{Xian_CVPR_2017}
Y.~Xian, B.~Schiele, and Z.~Akata.
\newblock Zero-shot learning - the good, the bad and the ugly.
\newblock In {\em IEEE Computer Vision and Pattern Recognition (CVPR)}, 2017.

\bibitem{Xu_Matrix_CVPR_2017}
X.~Xu, F.~Shen, Y.~Yang, D.~Zhang, H.~T. Shen, and J.~Song.
\newblock Matrix tri-factorization with manifold regularizations for zero-shot
  learning.
\newblock In {\em Proc. of CVPR}, 2017.

\bibitem{Ye_DSRL2017_CVPR}
M.~Ye and Y.~Guo.
\newblock Zero-shot classification with discriminative semantic representation
  learning.
\newblock In {\em The IEEE Conference on Computer Vision and Pattern
  Recognition (CVPR)}, July 2017.

\bibitem{Ying_JMLR_2012}
Y.~Ying and P.~Li.
\newblock Distance metric learning with eigenvalue optimization.
\newblock {\em J. Mach. Learn. Res.}, 13(1):1--26, Jan. 2012.

\bibitem{Yu_CVPR_2013}
F.~X. Yu, L.~Cao, R.~S. Feris, J.~R. Smith, and S.~F. Chang.
\newblock Designing category-level attributes for discriminative visual
  recognition.
\newblock In {\em Computer Vision and Pattern Recognition (CVPR), 2013 IEEE
  Conference on}, pages 771--778, June 2013.

\bibitem{Mubarak_2016_CVPR}
Y.~Zhang, B.~Gong, and M.~Shah.
\newblock Fast zero-shot image tagging.
\newblock In {\em The IEEE Conference on Computer Vision and Pattern
  Recognition (CVPR)}, June 2016.

\bibitem{Zhang_2015_ICCV}
Z.~Zhang and V.~Saligrama.
\newblock Zero-shot learning via semantic similarity embedding.
\newblock In {\em The IEEE International Conference on Computer Vision (ICCV)},
  December 2015.

\bibitem{Zhang_2016_CVPR}
Z.~Zhang and V.~Saligrama.
\newblock Zero-shot learning via joint latent similarity embedding.
\newblock In {\em The IEEE Conference on Computer Vision and Pattern
  Recognition (CVPR)}, June 2016.

\end{thebibliography}
}

\end{document}